\documentclass[runningheads]{llncs}

\usepackage{eccv}

\usepackage{eccvabbrv}

\usepackage{graphicx}
\usepackage{booktabs}

\usepackage[accsupp]{axessibility}

\usepackage[pagebackref,breaklinks,colorlinks,citecolor=eccvblue]{hyperref}

\usepackage{orcidlink}

\usepackage{mathtools}
\usepackage{array}
\usepackage{caption}
\usepackage{subcaption}
\usepackage{listings}
\usepackage{enumitem}
\usepackage{amsmath}
\usepackage{amssymb}
\usepackage[normalem]{ulem}
\usepackage{afterpage}
\usepackage{pifont} 
\usepackage{tikzsymbols}
\usepackage{wrapfig}
\usepackage{lipsum}

\newcommand{\cmark}{\ding{51}}%
\newcommand{\xmark}{\ding{55}}%
\newcommand{\clP}{\mathcal{P}}
\newcommand{\clN}{\mathcal{N}}
\newcommand{\clB}{\mathcal{B}}

\lstnewenvironment{myverbatim}[2][]{%
  \lstset{
    basicstyle=\ttfamily\scriptsize,
    frame=tb,
    label={lst:label},
    title=#2,
    #1
  }%
}{}

\begin{document}

\title{Open Vocabulary Multi-Label Video Classification}

\author{Rohit Gupta\inst{1}\orcidlink{0000-0002-9068-7429} \and Mamshad Nayeem Rizve\inst{2} \and
Jayakrishnan Unnikrishnan\inst{2} \and Ashish Tawari\inst{2} \and Son Tran\inst{2} \and 
Mubarak Shah\inst{1,2} \and \\ Benjamin Yao\inst{2} \and Trishul Chilimbi\inst{2}}

\authorrunning{R.~Gupta et al.}

\institute{Center for Research in Computer Vision, University of Central Florida \\
\email{rohit.gupta@ucf.edu, shah@crcv.ucf.edu} \and
Amazon \\ \email{\{jayunn,atawari,sontran\}@amazon.com}}

\maketitle

\begin{abstract}
Pre-trained vision-language models (VLMs) have enabled significant progress in open vocabulary computer vision tasks such as image classification, object detection and image segmentation. Some recent works have focused on extending VLMs to open vocabulary {\em single label} action classification in videos. However, previous methods fall short in holistic video understanding which requires the ability to  {\em simultaneously recognize multiple actions and entities} e.g., {\em objects} in the video in an open vocabulary setting. We formulate this problem as open vocabulary {\em multi-label} video classification and propose a method to adapt a pre-trained VLM such as CLIP to solve this task. We leverage large language models (LLMs) to provide semantic guidance to the VLM about class labels to improve its open vocabulary performance with two key contributions. First, we propose an end-to-end trainable architecture that \textit{learns} to prompt an LLM to generate \textit{soft attributes} for the CLIP text-encoder to enable it to recognize novel classes. 
Second, we integrate a temporal modeling module into CLIP's vision encoder to effectively model the spatio-temporal dynamics of video concepts as well as propose a novel regularized finetuning technique to ensure strong open vocabulary classification performance in the video domain. Our extensive experimentation showcases the efficacy of our approach on multiple benchmark datasets.

\end{abstract}

\section{Introduction}
\label{sec:intro}

\begin{figure}[t]
    \centering
    \includegraphics[width=0.85\linewidth]{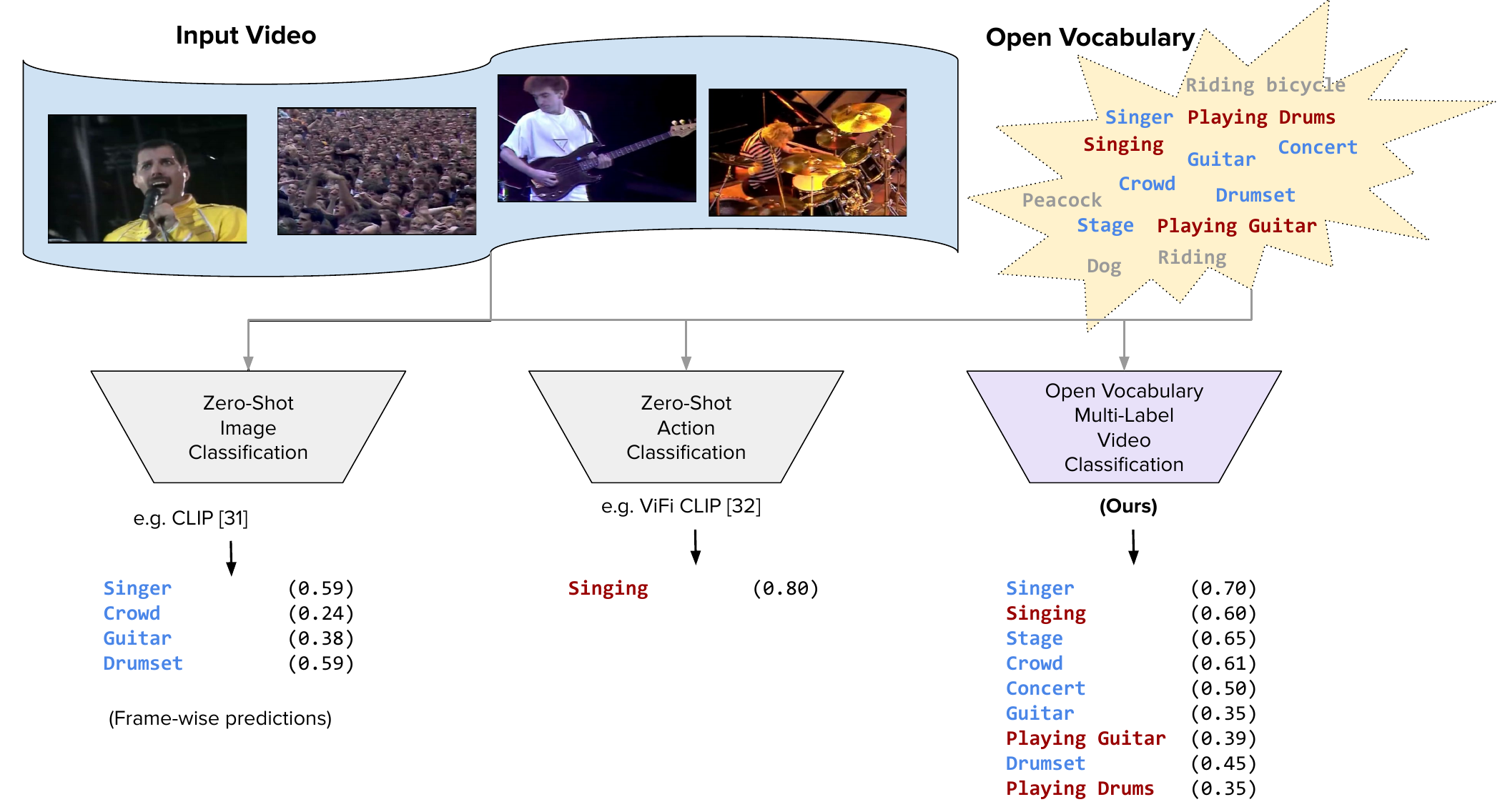}
    \caption{Our task: recognize multiple open-vocabulary classes in videos at inference from an open vocabulary, including \textcolor{blue}{\textbf{entities}} (in blue) such as objects and scenes, and \textcolor{red}{\textbf{actions}} (in red). Prior zero-shot {\em image} classification approaches (e.g. CLIP) can label the salient entity in each frame (left), while zero shot action-recognition approaches (e.g. ViFi-CLIP) (middle) can classify video-level {\em action}. Our method (right) can recognize all action or entity classes present in the video.}
    \label{fig:enter-label}
\end{figure}

Video classification is a critical challenge in computer vision. Its goal involves recognizing a diverse array of concepts depicted in a video, which may include primarily static entities such as objects and scenes, as well as dynamic actions. In the classic setting, the vocabulary of all possible classes of interest is known in advance, and the model is trained in a supervised manner using a labeled dataset. The labor-intensive process of manual annotation often results in video datasets that are narrowly focused, such as those limited to specific sports or simple activities, which restricts the breadth of concepts models can learn. As video applications are becoming widespread, there is an increasing need for developing video models that can recognize a broad range of concepts. This necessitates the development of open-vocabulary approaches for video classification that can recognize a diverse range of video concepts, including those that were not part of the class vocabulary present in the labeled training dataset. In this work, to solve this problem, we aim to leverage recent advances in vision-language models (VLM), which are trained to align visual and textual data and large language models (LLM), which have been demonstrated to possess a rich understanding of the world due to large scale pre-training.

VLMs that are trained for image-language alignment on image-text pairs at large-scale exhibit remarkable zero-shot visual recognition performance across a diverse set of tasks~\cite{radford2021learning, jia21b, Li_2022_CVPR, owlvit, Swetha_2023_ICCV, Swetha_icip, swetha2025implicitqa, swetha2025timelogic}. For classification tasks, the pretrained text encoder acts as a ``\textit{label encoder}'' for the class labels, mapping them into the same representation space as the the visual embedding of the image, which allows for the image to be classified by ranking the labels in order of representation similarity. Since VLMs are primarily trained to rank, picking the top match usually results in excellent performance on single label zero-shot classification. However, merely ranking the labels is insufficient to achieve open-vocabulary multi-label classification, and as a result, specialized methods~\cite{sun2022dualcoop} have been developed for multi-label image classification using VLMs. VLM based zero-shot image classification has been further improved by LLM-based prompting, in which LLMs are used to generate descriptive class definitions which are then used as prompts to the VLM text encoder during inference~\cite{pratt2023does, menon2023visual}. Other works have adapted VLMs for zero-shot \textit{single-label} action classification in videos \cite{wasim2023vita, weng2023open}. A typical video, however, contains multiple concepts including objects, actions, scenes, events, etc. Our goal is to develop a video classifier that can simultaneously recognize the presence of these different concepts from any vocabulary specified at inference.  We refer to this task as open-vocabulary multi-label video classification. An illustration of the subtle differences between the prior tasks and our proposed task is presented in Fig.~\ref{fig:enter-label}.

Open-vocabulary multi-label video classification presents two unique challenges. First, unlike the single label open-vocabulary setting, in the multi-label setting, identifying relevant concepts in a video based on VLM similarity score performs poorly, as VLM similarity scores for different types of concepts (e.g., actions, objects) often fall within different ranges, even with LLM-guided prompting (see Section~\ref{subsec:thresholds}).Addressing this issue requires end-to-end finetuning of the VLM with LLM guidance on datasets with diverse video concepts. However, performing end-to-end finetuning while simultaneously incorporating the benefits of LLM-guided prompting, presents additional technical challenges, particularly in backpropagating gradients through the VLMs text encoder's tokenizer. Second, adapting pretrained image-language models to recognize video concepts while retaining strong zero-shot performance is challenging, as the image-language pretraining datasets are orders of magnitude larger than the largest labeled video datasets. VLM adaptation tends to overfit to the video data easily, thus losing their generalization capability~\cite{wasim2023vita}.

In this work, we attempt to address these challenges for open vocabulary multi-label video classification. First, to simultaneously recognize multiple video concepts in open-domain, we finetune the VLM in an end-to-end manner with LLM guidance to ensure proper ranking between different types of video concepts.  Particularly, we extend the LLM-guided prompting approach for open-vocabulary classification by introducing learnable prefixes for prompting the LLM. To directly utilize the LLM output representations in the VLM's text encoder for end-to-end training, we propose a prompt transformer. This prompt transformer, in conjunction with the learnable prefixes, not only facilitates the integration of the LLM's world knowledge into the VLM's text representation but also aids in mitigating the discrepancy between VLM scores for various types of concepts. Second, for the effective recognition of temporally varying video concepts, we integrate an auxiliary temporal modeling branch with the VLM's vision encoder~\cite{liu2023revisiting} and introduce a novel regularization penalty to retain the VLM's zero-shot performance during the adaptation to videos. Through extensive experimentation and analysis we demonstrate that our proposed approach significantly outperforms baseline solutions obtained by extending prior work on single label open vocabulary image and video classification.

\noindent In summary, in this work we make three major technical contributions:
\begin{itemize}[leftmargin=*]
  \item A novel end-to-end trainable approach to learn a strong label encoder for open vocabulary multi-label video classification by adapting an LLM to learn to prompt a VLM's text encoder.%
  \item An approach for enhancing pre-trained VLM image encoders with temporal modeling capability while also retaining strong open-vocabulary performance.
  \item Defining a new open vocabulary multi-label video classification benchmark using 5 datasets (2 closed, 3 open vocabulary) and benchmarking our approach against 6 strong baselines, significantly outperforming them.
\end{itemize}

\section{Related Works}

\noindent{\bf Vision-Language Representation Learning:} Recently, image-language models \cite{radford2021learning, jia2021scaling, yu2022coca, li2021align, yao2021filip, singh2022flava} have drawn huge attention because of their effectiveness in learning generic visual representations that are transferable to several downstream tasks like classification, retrieval, etc. This success can partly be attributed to the large-scale image-text datasets \cite{schuhmann2021laion, schuhmann2022laion, 10.1145/2812802, desai2021redcaps, sharma2018conceptual}. However, this is not the case for the video data. Therefore, to perform video-language pretraining, most recent works \cite{ni2022expanding, luo2022clip4clip, fang2021clip2video, gorti2022x, xue2022clip, Swetha_2023_ICCV} bootstrap from a pretrained image-language model and adapt them to video representation tasks. 

\noindent{\bf Open-Vocabulary Classification:} Vision-language models do not perform optimally in visual recognition tasks out of the box. To improve the zero-shot open-vocabulary visual recognition capability of the vision-language models, the recent works either resort to regularized finetuning \cite{weng2023open, wortsman2022robust} or prompt learning \cite{zhou2022learning, wasim2023vita, sun2022dualcoop}. However, these works either focus on solving single label tasks or only perform adaptation for image recognition tasks. In contrast to these works, we focus on open-vocabulary video classification task.

\noindent{\bf Large Language Models in Vision:} Recently, Large Language Models (LLM) have been utilized in vision tasks primarily because of their excellent in-context and zero-shot learning performance especially in commonsense reasoning tasks~\cite{kojima2022large}. To be particular, LLMs have been used to rewrite noisy ASR text~\cite{shvetsova2023howtocaption, Gupta_2023_CVPR}, to generate classification labels by mapping textual description to a predefined task list~\cite{lin2022learning}, extracting verbs from textual descriptions to improve action understanding~\cite{lin2023match}. LLMs have also been used to initialize textual encoders for vision language encoder~\cite{zhao2023lavila}. In some recent works, vision representations have been aligned with LLM input space to solve multitude of vision-language tasks~\cite{pmlr-v162-li22n, NEURIPS2022_00226294, Swetha_xformer2024}. These multi-modal models have also been used to generate captions for images which are then utilized to train better vision-language models~\cite{fan2023improving, Swetha_safellava}.

\noindent{\textbf{Prompting Vision-Language Models with LLMs}:} Some prior works have leveraged LLMs to augment the open vocabulary understanding capabilites of Vision-Language models. Classification By Description~\cite{menon2023visual}, CuPL~\cite{pratt2023does},  LLM guided concept bottleneck~\cite{Yang_2023_CVPR} all utilize LLM generated class descriptors to prompt CLIP to improve image classification, and make it more interpretable. WaffleCLIP~\cite{Roth_2023_ICCV} demonstrated that LLM generated higher level concepts to describe datasets can also improve classification. LLM based approaches have also been extended to object detection~\cite{Kaul23} and point cloud understanding~\cite{Zhu_2023_ICCV}

\noindent{\textbf{Parameter efficient fine-tuning of LLMs}:} As full parameter finetuning of LLMs is not practical, quite a few parameter efficient finetuning methods have been developed. The lightweight Learnable Soft prompts~\cite{lester-etal-2021-power} is the approach we utilize to guide the LLM in our method. Other approaches such as LoRA~\cite{hu2022lora}, Prefix-Tuning~\cite{li-liang-2021-prefix} and P-Tuning~\cite{LIU2023} could also be adopted in principle.

\section{Method}
\label{sec:method}

\begingroup
\setlength{\intextsep}{5pt}%
\setlength{\columnsep}{10pt}%
Our proposal to boost the open vocabulary multi-label video classification capabilities of CLIP consists of two key parts: first, an end-to-end trainable label

\begin{wrapfigure}{r}{0.55\linewidth}
    \centering
    \includegraphics[width=\linewidth]{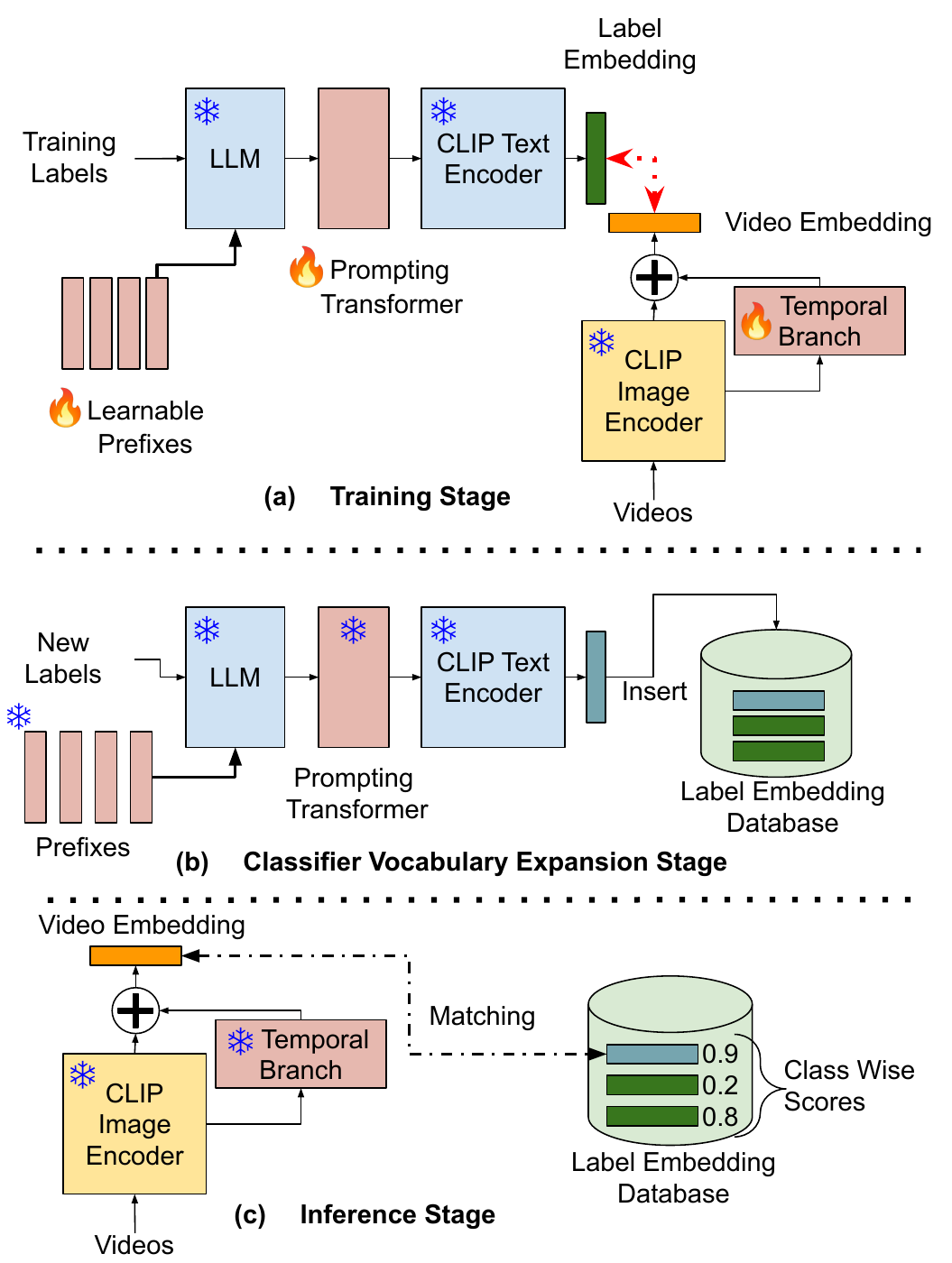} 
    \caption{Our open vocabulary classification method includes three stages of operation. During the \textbf{(a) training stage}, we train our label and video encoders on closed set training labels. New class labels can be added to the vocabulary after training by employing the \textbf{(b) classifier vocabulary expansion stage}. During the \textbf{(c) inference stage} video embeddings are computed and matched with the label embeddings database to get the classification scores.}
    \label{fig:overview}
\end{wrapfigure}

\noindent encoder leveraging an LLM for strong open vocabulary capabilities, and the CLIP text encoder for visual alignment. The label encoder consists of a frozen LLM guided with learnable prefixes, whose outputs are mapped to the frozen CLIP text encoder using a learnable prompting transformer. Second, we enhance the CLIP image encoder's capabilities for understanding temporal dynamics.  This is achieved by using a lightweight temporal modeling branch to enhance the CLIP image encoder. The details of our approach are illustrated in Figure~\ref{fig:method}. Each part of our method is discussed in detail in this section.

\noindent An overview of our approach  is provided in Fig.~\ref{fig:overview}. Our method has three broad stages.  Firstly, during the training phase, both the label encoder and video encoder are trained simultaneously, as shown in (a). 

\noindent   Secondly, during the classifier vocabulary expansion stage, embeddings for class labels are calculated and saved into a label embedding database, as shown in (b). This vocabulary can be extended at any point after training, thus allowing our method to be used in the open vocabulary setting. Finally, during the inference stage, video features are computed and compared against label embeddings from the database. As the label embeddings are pre-computed, the computational overhead during inference over standard CLIP models is minimal.

\endgroup

\subsection{LLMs for Semantic Enrichment of Label Embeddings}
\label{sec:prompting}

Pretrained VLMs have strong open-vocabulary image classification performance owing to their large scale image-text pretraining. LLMs on the other hand are trained on even larger scale of text data with training objectives that enable a rich semantic understanding. As a result, compared to LLMs, VLMs have a more limited semantic understanding of natural language and have difficulty understanding concepts such as the relationship between different class labels. For instance, prior works have found that while VLMs are able to classify ImageNet images at a fine-grained level, their accuracy at higher levels of the class hierarchy is poor~\cite{xu2023challenges}. E.g., while they may recognize an image of a \texttt{lion}, they are unable to understand that a \texttt{lion} is also a \texttt{feline}, a \texttt{mammal} and an \texttt{animal}. However, an LLM is very effective in comprehending the hierarchical relationship between these labels, among many other capabilities, due to the large-scale training. Hence, in this work, we utilize an LLM to generate complementary information about the class labels that can be utilized to improve the vision-language alignment of VLM leading to better open-vocabulary classification performance.

\begin{figure}[!h]
    \centering
    \includegraphics[width=0.85\textwidth]{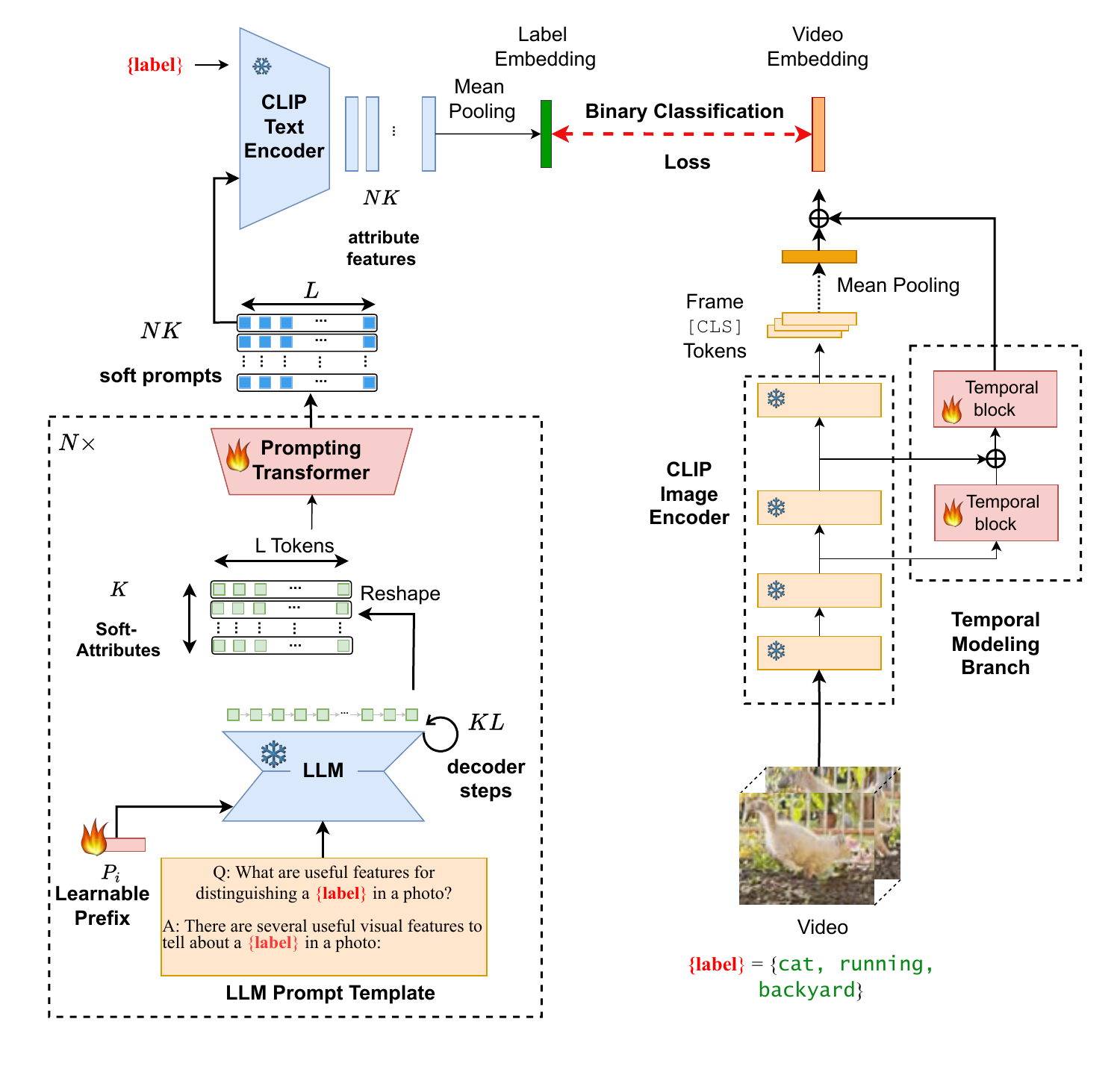} 
    \caption{Our end-to-end trainable system for open vocabulary video classification. The class labels are used by the LLM to generate useful class attributes for the CLIP text encoder which provides a visually aligned label embedding. The learnable input prompts to the LLM guide it to generate soft-attributes useful for video classification. Prompting transformer learns to map from the LLM output space to the CLIP input space. To add video understanding to CLIP's vision encoder we add additional spatio-temporal modeling layers. Details about each component are in Section~\ref{sec:method}.}
    \label{fig:method}
\end{figure}

\noindent{\textbf{Fixed LLM prompting.}}

As a baseline, we first develop an LLM-based prompting method to adapt CLIP for open vocabulary video classification, extending the approach of Menon \& Vondrick~\cite{menon2023visual} for open vocabulary image classification. We design a prompt template (see implementation details in Section K of Supplementary material) for an encoder-decoder based LLM with the class name and a question asking it to generate useful features for visually distinguishing that class. The LLM output is then parsed into a list of textual descriptions, hereafter referred to as attributes. These attributes along with the class name is used to prompt the CLIP text encoder and the resultant text-embeddings are mean-pooled to obtain an attribute enriched text-embedding for the class. We mean-pool CLIP vision embedding of video frames to generate video embeddings and then perform open vocabulary classification by matching video embeddings with the text-embeddings of different classes. However, this simple baseline has a crucial weakness: it is not trainable, as there is a need to de-tokenize and tokenize the LLM text output to CLIP, which doesn't allow the flow of gradients. As a result, the process of generating class attributes by prompting an LLM cannot be improved by training on a labeled video dataset. To remedy this, we propose an end-to-end trainable architecture that integrates the LLM with the CLIP text encoder.

\noindent{\textbf{End-to-end learnable LLM prompting.}}
Our proposed architecture, summarized in Figure~\ref{fig:method}, incorporates a learnable prompting framework with the frozen LLM to generate the inputs to a frozen CLIP text-encoder. The learnable components on the text side are limited to $N$ learnable prefixes/vectors to the prompt template used for querying the LLM and a prompt transformer that transforms the sequence of tokens from the LLM to input soft prompts for the CLIP text-encoder. In our implementation we avoid the discrete and non-differentiable operations like detokenization at the LLM decoder output and re-tokenization at the CLIP text-encoder input. This allows the prompt transformer to directly connect the LLM output semantic space to the CLIP input semantic space, ensuring that the entire text model is differentiable and therefore end-to-end learnable. We describe the details below.

For each of the $N$ LLM prefixes, we first construct an input sequence to the LLM comprising of the prefix followed by a fixed prompt template that contains the label name $\ell$ from the input data sample (see Figure~\ref{fig:method}). The text input to the LLM is mapped using a tokenizer and the LLM's embedding layer into a sequence of tokens \(I \in \mathbb{R}^{M \times d}\), where \(M\) and \(d\) represent number of input tokens and the dimension of the embedding space respectively. All our $N$ learnable prefixes are $d$-dimensional vectors. We use $P_i \in \mathbb{R}^d$ to represent the $i$-th prefix. Each prefix is concatenated with the tokens of the prompt template, yielding a unified sequence of tokens \([P_i; I] \in \mathbb{R}^{(1+M) \times d}\). This combined sequence is then processed through the frozen encoder-decoder layers of the LLM. For each prefix, we run $KL$ decoding iterations of the LLM decoder to generate a sequence of $KL$ decoded tokens. These tokens represents useful class features in the LLM's semantic space. As the prompt templace specifically asks LLM to output features as a list, we split the sequence of tokens evenly into $K$ subsequences of $L$ tokens each. Due to the nature of LLM prompt template, which prompts the LLM to output useful visual features, we refer to the subsequences of tokens as \textit{soft attributes}, as they are sets of continuous vectors instead of discrete attributes in natural language. The $K$ soft attributes are then individually processed by the prompt transformer to generate $K$ \textit{soft prompts} to the CLIP text encoder. Repeating this operation for each of the $N$ LLM prefixes, we get $NK$ soft prompts. Each soft-prompt is then concatenated with the tokenized label embedding and then processed by the frozen CLIP text-encoder to generate an attribute feature. All $NK$ resulting features are mean-pooled and normalized to obtain the final label embedding $f_t(\ell)$, where $\ell$ is the label name. 

\subsection{Regularized Parallel Temporal Modeling}

\begingroup
\setlength{\intextsep}{5pt}%
\setlength{\columnsep}{10pt}%

\begin{wrapfigure}{r}{0.5\linewidth}
     \centering
         \includegraphics[width=\linewidth]{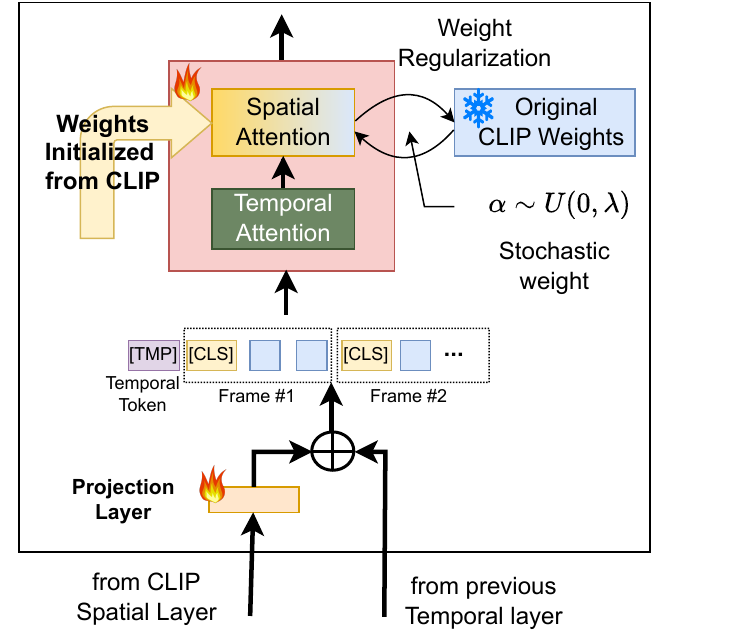}
        \caption{Our Temporal Block projects frame patch tokens from the CLIP image encoder, fusing them with previous block's temporal branch tokens. The temporal token (TMP) incorporates all frames' CLS tokens. Divided Space-Time attention layers form the core of the block. Spatial attention layers are initialized from CLIP weights and regularized using stochastic weight averaging. }
        \label{fig:temporalstruct}
\end{wrapfigure}

The simple image transformer of the CLIP vision encoder does not model the temporal dynamics of the video, which is essential to enhance the ability to recognize entities in videos. We enhance our vision encoder by adding a parallel temporal modeling branch to the last $T$ layers of the CLIP vision encoder as illustrated in Figure~\ref{fig:method}. We freeze the CLIP vision branch and train only the newly added temporal layers. Each block in the newly added branch consists of a spatial attention layer initialized from their corresponding CLIP weights, and a temporal attention layer that is randomly initialized.
All frames of the video are processed independently by the CLIP vision backbone. Then, at the first temporal modeling block, the temporal token TMP is created by averaging the CLS tokens across frames, meanwhile learnable spatial and temporal positional embeddings are added to each patch token. The $t$th temporal modeling block takes in as input a weighted combination of the previous temporal modeling block and the corresponding layer in the CLIP backbone. Symbolically if ${V_t}^{\intercal}$ represents the patch tokens from the temporal block at the $t$th block and ${V_t}$ represents the patch tokens for the corresponding CLIP layers, the patch tokens for the $t$th block ${V_t}^{\intercal} = {V}^{\intercal}_{t-1}  + Proj_{spatial}(V_s)$, where $V_s$ is the corresponding token from the CLIP backbone. After that the tokens are processed by the divided space-time attention layer and spatial layer. The overall video embedding is generated at the end by mean pooling the final TMP token and the CLS tokens for each frame from the CLIP backbone.

\endgroup

\noindent A key drawback of adding a deep parallel branch is that while it improves performance in the finetuned setting, it diminishes CLIP's zero shot performance. In order to overcome this limitation in the zero-shot setting, we propose to use weight regularization on the spatial attention layers. The weight regularization operation tries to match predictions made by the current set of finetuned weights and a randomly weighted average of the current weights and the original CLIP weights. Symbolically, at each iteration we use weights $\theta$ which are a stochastic weighted average of the finetuned and the original frozen weights as given by:

\[
\theta = \alpha \theta_{\text{ft}} + (1 - \alpha) \theta_{\text{frozen}}, \quad \text{where} \quad \alpha \sim U(0, \lambda)
\]

\noindent We observe that an empirical value of $\lambda = 0.5$ works well across datasets. \\

\noindent \textbf{Relationship to prior works:} Prior temporal modeling approaches split into two camps: improving finetuned performance or preserving CLIP's zero-shot abilities. STAN~\cite{liu2023revisiting} is a state-of-the-art finetuning focused approach which adds a parallel temporal modeling branch which is able to leverage both high-level and low-level features from the CLIP vision encoder. The divided space-time attention in STAN's temporal block is inspired by the TimeSformer~\cite{gberta_2021_ICML} architecture. We differ from STAN in order to achieve stronger open vocabulary performance in a number of ways. Firstly, unlike STAN we do not finetune the CLIP image encoder, only the added parallel branch. Additionally, we propose stochastic weight regularization to prevent the temporal branch from drifting too far from the original CLIP feature space. Our ablation experiments (Table~\ref{tab:side-by-side}) demonstrate that both these choices significantly improve our performance. Some form of weight regularization has been explored by methods such as Open-VCLIP\cite{weng2023transforming}. However, Open V-CLIP takes in only 3 frames at a time, and its temporal modeling range is limited. Our approach is able to achieve the best of both approaches.

\subsection{Training objective}

We train our model on multi-label video datasets. To construct a batch, we first sample a set $\clB$ of $B$ videos from the dataset. Each video $v$ is associated with a set of positive class labels which we denote $\clP(v)$. In addition to the positive labels, we augment the batch with random negative class labels for each video. To obtain the negatives, we first identify the set $\clP_\clB := \cup_{v \in \clB} \clP(v)$ of distinct classes among all the positive labels from all the videos in the batch. We then sample a random set $\clN_\clB$ of $4B - |\clP_\clB|$ classes from the rest of the class vocabulary of the dataset. For each video $v$, we then choose all non-positive classes from $\clP_\clB \cup \clN_\clB$ as negative labels. We use $\clN(v) := (\clP_\clB \cup \clN_\clB) \setminus \clP(v)$ to denote the set of all negative labels for $v$. The resulting positive and negative video-label pairs are then treated as training samples for binary classification. Each training sample is a label, video pair $(\ell, v)$ where $\ell$ represents the label name and $v$ the video. With this batch construction, the total number of training samples in a batch could be variable, but the total number of videos is fixed at $B$ and the total number of classes present among the samples is fixed at $4B$.

\begin{wrapfigure}{r}{0.45\linewidth}
     \centering
         \includegraphics[width=\linewidth]{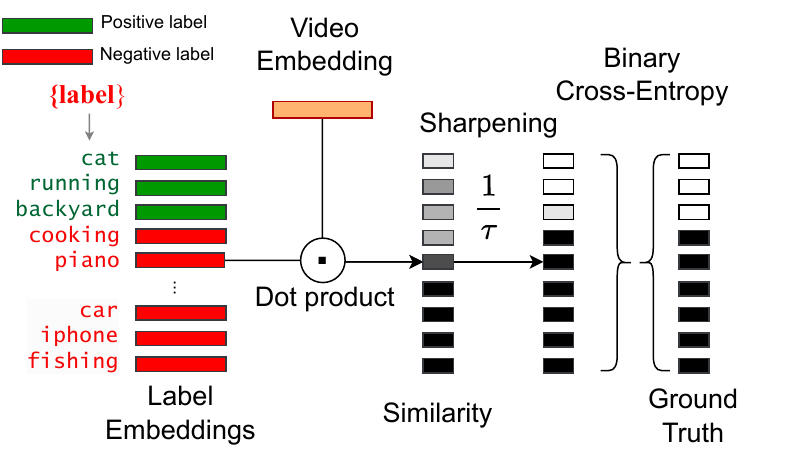}
        \caption{Our training objective applies binary cross-entropy to predicted Video-Label feature similarities sharpened by a temperature-scaled sigmoid.}
        \label{fig:lossfunc}
\end{wrapfigure}

\noindent  For each training data sample $(\ell, v)$, the text and video encoders respectively generate a unit-norm text embedding $f_t(\ell) \in \mathbb{R}^D$ and a unit-norm  video embedding $f_v(v) \in \mathbb{R}^D$ where $D$ is the common embedding dimension. The score $s(\ell,v)$ for this data sample is given by the inner product:
\begin{equation}
s(\ell, v) = (f_t(\ell))^{\intercal}f_v(v). \label{eqn:score}
\end{equation}

\noindent The model is trained with a weighted binary cross entropy loss: 
\begin{align}
\hspace{-0.1in} \mathcal{L}(\clB) \hspace{-0.01in} = \hspace{-0.01in} -\sum_{v \in \clB} \hspace{-0.01in} \left[ \hspace{-0.01in} \sum_{\ell \in \clP(v)} \hspace{-0.05in} \log p(\ell,v) + \hspace{-0.01in} w \hspace{-0.05in} \sum_{\ell \in \clN(v)}\hspace{-0.05in} \log(1-p(\ell,v))\right] \label{eqn:bceloss}
\end{align}
\normalsize 
where $p(\ell,v) := \sigma\left(\frac{s(\ell,v)}{\tau}\right)$, $\tau$ is a temperature parameter, $\sigma(.)$ is the sigmoid function and $w > 0$ is a weight hyperparameter.

\section{Experiments and Results}

\subsection{Datasets}

In order to train our model on a wide range of concepts, we train it on a mix of YouTube8M~\cite{abuelhaija2016youtube8m} and Kinetics-400~\cite{kay2017kinetics}. YouTube-8M (YT-8M) is primarily labeled with entities, whereas Kinetics-400 (K400) is labeled only with actions. As YT-8M is composed of a random sample of the \emph{whole} of YouTube, a significant portion of video its samples are from video game streams. Many of these videos are just tagged with the video game title as the label irrespective of the actual entities and actions present in the video. A significant portion of its label vocabulary is dedicated to video game titles (771 out of 3862 total) are dedicated to video game titles. We observe training instabilities due to this, and to fix this issue, we remove video game titles from the \emph{training} vocabulary. We also remove some of the least frequently occurring labels during training to reach a training vocabulary of 2429 classes. This lead to stabilized training without further issues. Note that for closed set evaluation on YT-8M, we use the YouTube-8M Segments validation set, which contains human verified labels for 1000 classes. K400 on the other hand is widely used, and has a clean vocabulary and high quality human verified action labels. From the training perspective, the only limitation is the absence of entity (object, scenes etc.) labels. 

For evaluation, we use 3 open-vocabulary test datasets: TAO (Tracking Any Object)~\cite{dave2020tao}, ActivityNet~\cite{activitynet} and RareAct\cite{miech20rareact}.
TAO and ActivityNet are exclusively object and action focused datasets respectively. RareAct has object, action labels, and focuses on unusual object-action combination. TAO dataset was developed for evaluating object trackers, however analogous to how object detection datasets like MS-COCO are used for evaluating multi-label image classification methods such as DualCoOp~\cite{sun2022dualcoop}, we transform TAO into a multi-label video classification dataset by ignoring localization annotations. ActivityNet often has multiple actions in a given video. For our evaluation each annotation from RareAct provides three labels: object, action and the (unusual) object-action combination, this provides an interesting test of the open vocabulary capabilties. Altogether, these five datasets provide a comprehensive evaluation of a model's closed and open vocabulary video classification performance.

\subsection{Baselines}
As there are no prior works that explicitly address multilabel open vocabulary video classification, we extend the following methods from the zero shot image and action classification literature to our multi-label video classification setting for comparison. 

{\noindent \textbf{CoOp}\cite{zhou2022learning}:} Short for \textbf{Co}ntext \textbf{Op}timization, CoOp learns prompts for the CLIP text encoder as a lightweight adaptation technique for classification. 

{\noindent \textbf{DualCoOp}\cite{sun2022dualcoop}:} Is an extension of CoOP to the multi-label setting where both positive and negative prompts are learnt for the CLIP text encoder. For a given label, prediction is based on whether the positive or negative prompted version has higher similarity with the image feature. 

{\noindent \textbf{LLM + CLIP (Frozen)}:} This baseline was discussed in Section~\ref{sec:prompting} and is illustrated in Figure 8 of Supplementary.

{\noindent \textbf{ViFi-CLIP}~\cite{hanoonavificlip}:} In this baseline, there is no temporal modeling, and both the CLIP image and text encoder are finetuned on the training dataset.

As \texttt{CoOp} and \texttt{ViFi-CLIP} were developed for single label classification, to utilize them in our setting we replace their contrastive loss function with our multi-label classification loss. We also drop the region aggregation aspect of \texttt{DualCoOp} and only test the dual prompting architecture. We refer to these baselines as \textbf{CoOp}$^{\ast}$, \textbf{DualCoOp}$^{\ast}$ and \textbf{ViFi-CLIP}$^{\ast}$ to reflect these differences.

\begin{table*}
\centering
\scriptsize
\begin{tabular}{lccccc} 
\toprule
& \multicolumn{2}{c}{\textbf{Closed-Vocabulary}} & \multicolumn{3}{c}{\textbf{Open-Vocabulary}} \\
\cmidrule(lr){2-3} \cmidrule(lr){4-6}
\textbf{Method} & \multicolumn{1}{c}{YouTube-8M} & \multicolumn{1}{c}{Kinetics} & \multicolumn{1}{c}{\begin{tabular}[c]{@{}c@{}}TAO\\\tiny{(Entities)}\end{tabular}} & \multicolumn{1}{c}{\begin{tabular}[c]{@{}c@{}}ActivityNet\\\tiny{(Actions)}\end{tabular}} & \multicolumn{1}{c}{\begin{tabular}[c]{@{}c@{}}RareAct\\\tiny{(Entities+Actions)}\end{tabular}} \\ 
\midrule
\multicolumn{3}{l}{\textit{Frozen CLIP-based Methods}} \\ 

CLIP with Class name Prompt   & 6.3  & 26.2 & 43.8  & 44.2 & 9.5 \\ 
CLIP with Prompt Templates & 6.8 & 30.5 & 46.0   & 45.9 & 11.4 \\ 
CLIP with Fixed LLM Prompts   & 6.9 & 30.6 &  50.2  & 46.8 & 11.5  \\ 
\midrule
\multicolumn{3}{l}{\textit{Trainable Baseline Methods}} \\
CoOp      & 2.7 & 17.8 & 35.0 & 28.8 & 3.5 \\ 
DualCoOp  & 8.3 & 23.9 & 47.1  & 33.0 & 7.6 \\
ViFi-CLIP & 3.4  & 10.9 & 58.3 & 17.2 & 4.1 \\
\midrule
\multicolumn{3}{l}{\textit{\textbf{Ours}}} \\

CLIP + Learnable LLM Prompts & 9.4 & 32.8 &  51.4 & 47.1 & 11.9 \\ 
 + Temporal Modeling & 14.8 & 42.0 & 63.8 &  47.1 & 12.4 \\ 
  + Synthetic Labels & \textbf{16.7} & \textbf{43.2} & \textbf{65.5} & \textbf{50.2} &  \textbf{13.2}\\

\bottomrule
\end{tabular}

\caption{AUPR scores for all methods on all datasets.}

\label{tab:aupr}
\end{table*}

\subsection{Metrics}

We report two sets of metrics to evaluate the performance of the models in the open vocabulary setting. Area Under Precision-Recall curve (AUPR) summarizes the overall classification performance across the entire precision-recall trade-off. Secondly, we report the Peak F1-Score for the model on each dataset, which is the F1-Score achieved by that model at the optimal threshold chosen by an oracle for that dataset. This metric captures the classification performance obtained by an open vocabulary classification method on a dataset, if the threshold alone could be tuned, e.g., by using a labeled validation set.

\subsection{Results}

AUPR scores for each method are presented in Table~\ref{tab:aupr} and Peak F1 scores are presented in Table~\ref{tab:peakf1}. Our learnable LLM prompting approach outperforms both frozen and trainable baselines across both closed-vocabulary and open-vocabulary classification tasks. Also notable is the enhancement in performance due to our temporal modeling approach (average gain of $\approx 5\%$ in AUPR). As previously mentioned in our discussion of the dataset, the training datasets YT-8M and K400 have certain limitations (in particular, YT-8M has few action labels and K400 has no object/entity labels). We present further improved results obtained by augmenting our training dataset with additional class labels obtained using a multimodal LLM (pipeline is detailed in Section I of the Supplementary Material). As seen from the last rows of Tables~\ref{tab:aupr} and~\ref{tab:peakf1}, this approach gives a further $\approx$ 2\% improvement, and is specially effective for actions.

 \begin{table*}

\centering
\scriptsize
\begin{tabular}{lccccc} 
\toprule
& \multicolumn{2}{c}{\textbf{Closed-Vocabulary}} & \multicolumn{3}{c}{\textbf{Open-Vocabulary}} \\
\cmidrule(lr){2-3} \cmidrule(lr){4-6}
\textbf{Method} & \multicolumn{1}{c}{YouTube-8M} & \multicolumn{1}{c}{Kinetics} & \multicolumn{1}{c}{\begin{tabular}[c]{@{}c@{}}TAO\\\tiny{(Entities)}\end{tabular}} & \multicolumn{1}{c}{\begin{tabular}[c]{@{}c@{}}ActivityNet\\\tiny{(Actions)}\end{tabular}} & \multicolumn{1}{c}{\begin{tabular}[c]{@{}c@{}}RareAct\\\tiny{(Entities + Actions)}\end{tabular}} \\ 
\midrule
\multicolumn{3}{l}{\textit{Frozen CLIP-based Methods}} \\ 

CLIP with Class name Prompt   & 14.9  & 34.2 & 44.6 &  47.1 & 17.6\\ 
CLIP with Prompt Templates & 20.8  & 36.8 & 49.2 &  50.1 & 20.5 \\ 
CLIP with Fixed LLM Prompts   & 21.6 & 37.3 & 50.2 &   51.4 & 19.8 \\ 
\midrule
\multicolumn{3}{l}{\textit{Trainable Baseline Methods}} \\
CoOp & 5.8 & 25.5 & 43.9 &  35.5 &  10.5 \\ 
DualCoOp & 16.2 & 33.2 & 49.0 &  40.5 & 15.0 \\
ViFi-CLIP & 5.2 & 19.3 & 54.4 &  24.7 & 9.6\\
\midrule
\multicolumn{3}{l}{\textit{\textbf{Ours}}} \\

CLIP + Learnable LLM Prompts & 23.6 & 42.4 &  52.8 &  51.1 & 22.6 \\ 
 + Temporal Modeling & 31.5 & 46.2 &  \textbf{59.6} &  52.6 & 24.3\\ 
  + Synthetic Labels & \textbf{32.7} & \textbf{46.6} &  56.6 &  \textbf{53.8} &  \textbf{25.1}\\

\bottomrule
\end{tabular}
\caption{Peak F1 scores for all methods on all datasets.}
\label{tab:peakf1}
\end{table*}

\subsection{Improved score calibration across datasets}
\label{subsec:thresholds}

\begin{figure*}[t]
\centering
\begin{subfigure}{0.45\linewidth}
    \includegraphics[width=\linewidth]{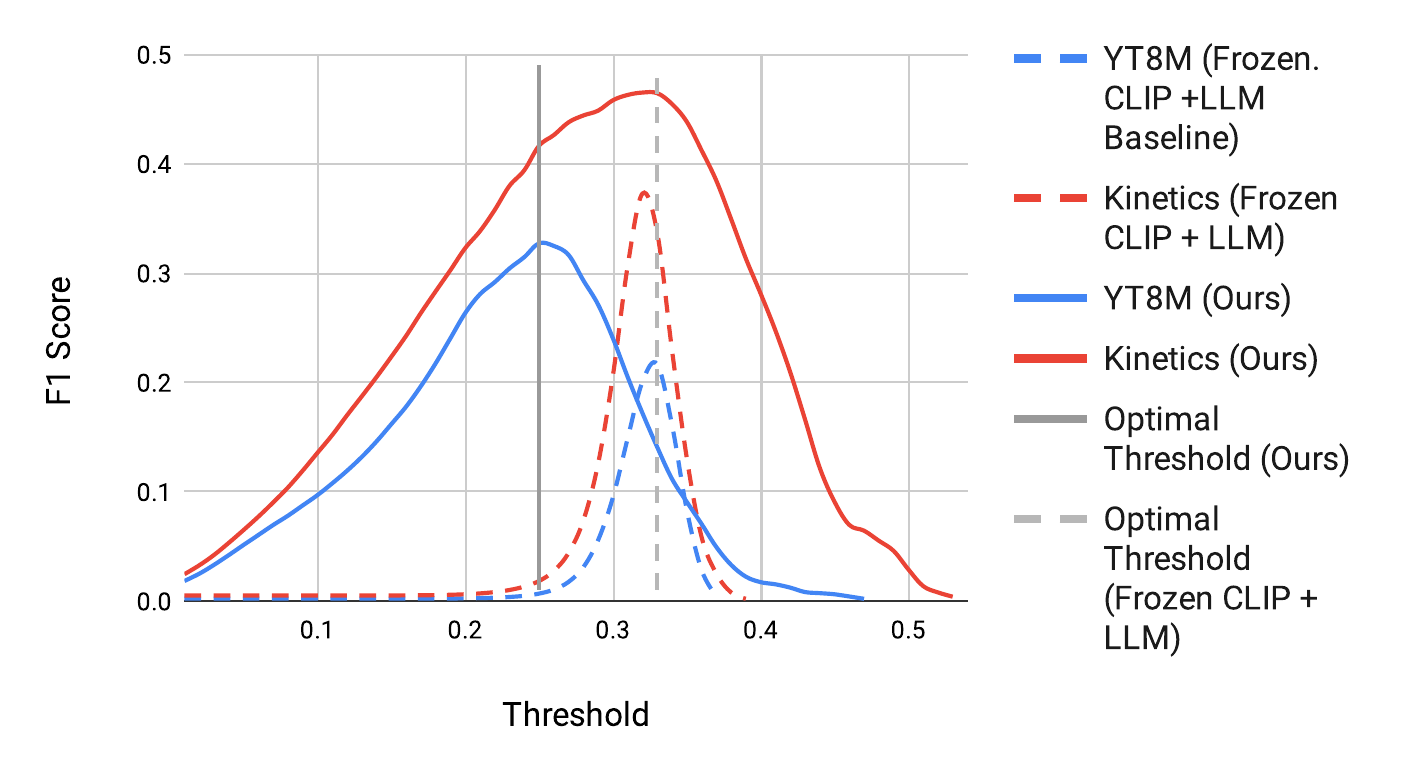}
    \caption{Closed Vocabulary Evaluation.}
    \label{fig:supcal}
\end{subfigure}
\hfill
\begin{subfigure}{0.45\linewidth}
    \includegraphics[width=\linewidth]{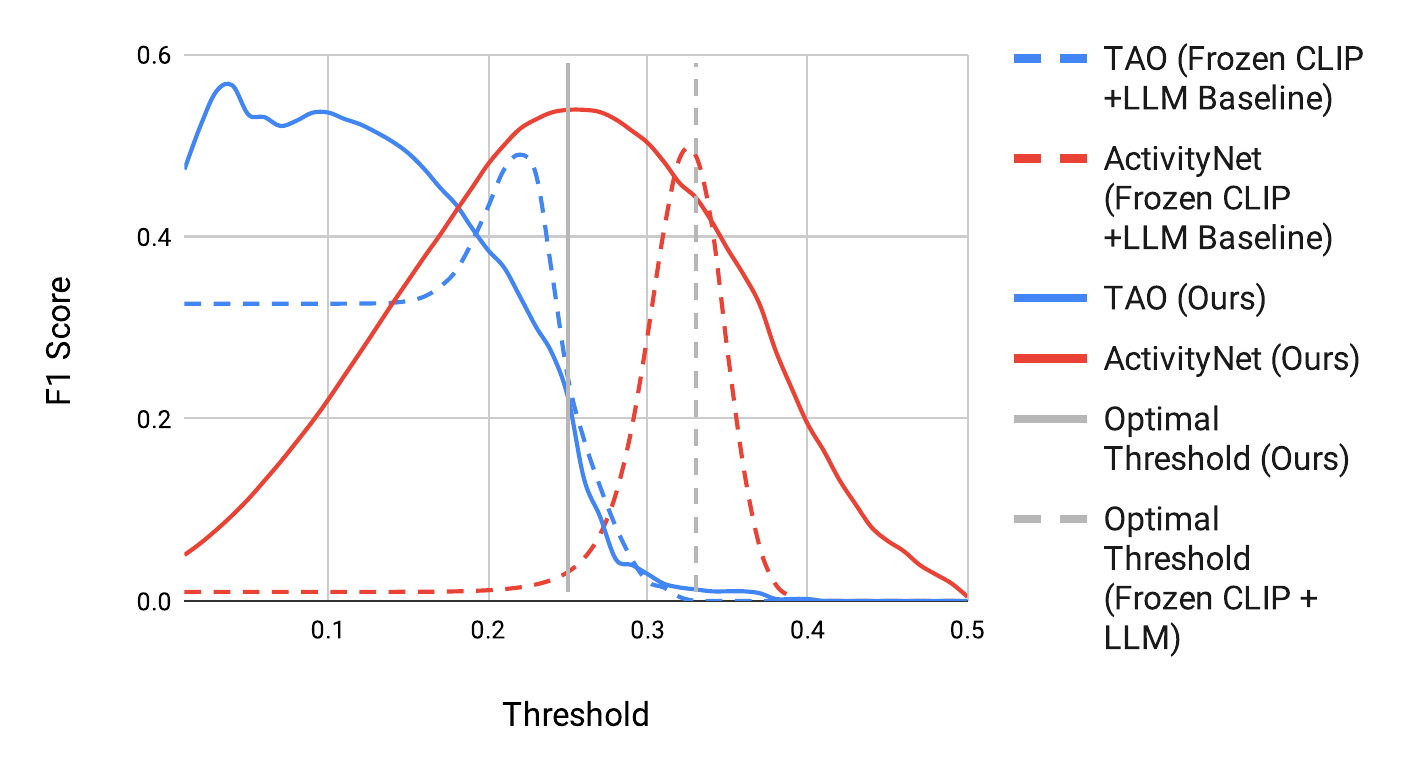}
    \caption{Open Vocabulary Evaluation.}
    \label{fig:zscal}
\end{subfigure}

\caption{F1-Scores at different thresholds for closed and open vocabulary evaluation datasets. Our end-to-end trained model can achieve better performance across datasets with a single threshold (labeled by gray vertical line) chosen on the supervised datasets.}

\label{fig:calibration}
\end{figure*}

In order to use a classifier in a truly open vocabulary setting in practice, the classification score needs to be well calibrated across different types of concepts. This would allow us to pre-select an optimal threshold to generate binary classification for all concepts. In Figure~\ref{fig:calibration} we demonstrate the robustness offered by our method in setting a threshold that works for a wide range of concepts, and the advantage offered by our method over the frozen CLIP + LLM baseline. These figures plot the F1 scores for multi-label classification across thresholds provided by both methods with Figure~\ref{fig:supcal} showing the numbers on the validation splits of the training datasets, and Figure~\ref{fig:zscal} showing the numbers in the open vocabulary setting on two diverse datasets. A reasonable choice for a classification threshold would be to pick the threshold that maximizes the minimum validation F1-score for the datasets used in training. The resulting choice of thresholds for our method and for frozen CLIP are shown in Figure~\ref{fig:supcal}. We see that for the baseline method the performance on TAO is poor, while that on ActivityNet is modest. However with our proposed method the F1 scores for both evaluation datasets are simultaneously high. This result empirically validates the versatility that is achieved by our learnable LLM-based prompting scheme over the baseline, which makes it possible to pre-select a threshold for using the solution as a black box classifier for all unseen concepts. 

\subsection{Ablations}

We scientifically ablate each part of our framework, reporting ablation evaluations (AUPR) on TAO (Objects) and ActivityNet (Actions).

\begin{table}[t]
    \centering
    \begin{subtable}[t]{0.34\textwidth}
        \centering
        \scriptsize
        \begin{tabular}{ccc} 
            \toprule
            \multicolumn{1}{c}{} & \multicolumn{2}{c}{\textbf{Open Vocabulary}} \\ 
            \midrule
            \multicolumn{1}{c}{\textbf{\# Blocks}} & \multicolumn{1}{l}{\textbf{Objects}} & \multicolumn{1}{l}{\textbf{Actions}} \\ 
            \midrule
            5 & 65.4 & 48.1 \\ 
            4 & \textbf{65.5} & \textbf{50.2} \\ 
            2 & 64.6 & 49.1 \\ 
            1 & 62.3 & 47.7 \\
            \bottomrule
        \end{tabular}
        \caption{Effect of Temporal Modeling}
        \label{table:temporalabl}
    \end{subtable}
    \hfill
    \begin{subtable}[t]{0.62\textwidth}
        \centering
        \scriptsize
        \begin{tabular}{ccccc} 
            \toprule
            \multicolumn{2}{c}{} & \multicolumn{3}{c}{\textbf{Open Vocabulary}} \\ 
            \midrule
            \multicolumn{1}{c}{\textbf{Spatial Reg.}} & \multicolumn{1}{c}{\textbf{Backbone}} & \multicolumn{1}{l}{\textbf{Objects}} & \multicolumn{1}{l}{\textbf{Actions}} & \multicolumn{1}{l}{\textbf{Geo. Mean}}\\ 
            \midrule
            \checkmark & {\large \ding{100}} &  56.6 & 53.8 & \textbf{55.2} \\
            \xmark & {\large \ding{100}} & 62.1 & 32.5 & 44.9 \\
            \xmark & {\Large \textcolor{orange}{\Fire}} & 42.7 & 28.3 & 34.8 \\
            \bottomrule
        \end{tabular}
        \caption{Effect of our Regularization}
        \label{table:temporalreg}
    \end{subtable}
    \caption{Ablations of Temporal Modeling Branch Architecture and Regularization}
    \label{tab:side-by-side}
\end{table}

\begin{table}[t]
\centering
\scriptsize
\begin{tabular}{ccccc} 
\toprule
\multicolumn{3}{c}{\textbf{Label Encoder}} & \multicolumn{2}{c}{\textbf{Open Vocabulary}} \\ 
\midrule
\textbf{LLM} & \textbf{PT} & \textbf{CLIP} & \textbf{Objects} & \multicolumn{1}{l}{\textbf{Actions}} \\ 
\midrule
 Learnable Prompt  & \cmark & \xmark & 25.2 & 8.5 \\
 Fixed  Prompt     & \xmark & \cmark & 62.5 & 41.3 \\
 Fixed  Prompt     & \cmark & \cmark & 64.9 & 48.1 \\
 Learnable Prompt  & \cmark & \cmark & \textbf{65.5} & \textbf{50.2} \\
 \xmark            & \xmark & \cmark  & 60.1 & 34.5 \\ 
\bottomrule
\end{tabular}

\caption{Ablating the Label Encoder. PT $\rightarrow$ Prompt Transformer}
\label{table:textablation}
\end{table}

\begin{itemize}[noitemsep,nolistsep]
    \item \textbf{Temporal Modeling Blocks:} We find that using more Blocks improves performance, until around 4. (Table~\ref{table:temporalabl})
    \item \textbf{Temporal Modeling Weight Regularization:} We find that using our weight regularization strategy for the spatial layers of the temporal branch prevents the model from overfitting to certain concepts. We also validate our choice of keeping the CLIP vision backbone frozen and find that unfreezing the backbone leads to severe over-fitting. (Table~\ref{table:temporalreg})
    \item \textbf{Label Encoder:} We demonstrate the effectiveness of combining an LLM with learnable prompts and CLIP's text encoder as the label encoder. First, we show that removing CLIP text encoder leads to a significant drop in performance. This demonstrates that the visual-text alignment learned by CLIP is essential. Next we demonstrate the benefits of learnable prompts over fixed prompts. We also provide results with the LLM removed (Table~\ref{table:textablation}).
\end{itemize}

\vspace{-0.5em}
\section{Conclusion}
\vspace{-0.5em}
We introduced the problem of open vocabulary multi-label video classification and proposed a solution leveraging LLMs and pre-trained VLMs. Through our extensive experiments we demonstrated strong performance on both actions and entities with a single model. Our proposed approach benefits from two key innovations. First, we proposed a method to prompt VLMs more effectively by utilizing the LLM output representations through a trainable prompt transformer and learnable LLM prompts. Second, we introduced a temporal modeling architecture and a regularized finetuning approach to improve the video understanding capability of the vision encoder while retaining strong open-vocabulary performance. Our ablations validate the efficacy of each of these contributions.

\appendix
\clearpage

\begin{center}
{\Large \textbf{Supplementary Material} for \textit{Open Vocabulary Multi-Label Video Classification}}
\end{center}

\renewcommand{\arraystretch}{1.2}

\section*{Overview}

This supplementary material is organized into the following sections:

\vspace{1mm}
\begin{itemize}

\item Section~\ref{sec:comparisonsup} Comparison of our results with Supervised State of the Art
\item Section~\ref{sec:comparisonopen} Comparison of our results with Single Label Open Vocabulary Baselines
\item Section~\ref{sec:comparisonllava} Comparison of our results with a Multi-Modal LLM

\item Section~\ref{sec:singlelabel}: Evaluation of our approach on Single Label Classification tasks
\item Section~\ref{sec:egocentric}: Evaluation of our approach on EgoCentric tasks
\item Section~\ref{sec:labablateextra}: Additional Ablations for Label Encoder
\item Section~\ref{sec:tempablateextra}: Additional Ablations for Temporal Encoder

\item Section~\ref{sec:computecost}: Inference and training costs of our approach
\item Section~\ref{sec:pipelinesupp}: Further details about the Synthetic Label Pipeline
\item Section~\ref{sec:baselinessupp}: Further details about all the Baselines reported
\item Section~\ref{sec:implementsupp}: Implementation details
\item Section~\ref{sec:qualitative}: Qualitative Results
\end{itemize}

\section{Comparison with Supervised SOTA}
\label{sec:comparisonsup}

In order to provide some additional context for our results, we also evaluate some existing state of the art baselines on our downstream datasets.

The best ActivityNet trained model with public weights is ASM-Loc (He et al.~\cite{he2022asm}, CVPR 2022). Our open vocabulary classifier comes within 10\% Peak F1-Score of this supervised model despite not being trained on any ActivityNet data. We also provide results finetuning ViFi CLIP on downstream datasets, which diminishes open vocabulary generalization capabilities. Our single model is competitive across both datasets.

\begin{table}[h]
    \vspace{-1em}
    \caption{\textbf{Comparing with Supervised Results} (Peak F1-Score)} \label{tab:sup}
    \vspace{-1em}
    \scriptsize
    \centering
    \begin{tabular}[t]{l|cc|c}
        \toprule
        \textbf{Method} & TAO & ActivityNet & Geometric Mean \\
        \midrule
        {\bf Ours (Zero-Shot)} & 56.6 & 53.8 & \textbf{55.2} \\
        \midrule
        \multicolumn{3}{l}{\textbf{Supervised Methods}} \\
        \texttt{ASM-Loc} \tiny{\texttt{CVPR 2022}} & - & 63.1 & - \\
        \texttt{ViFi-CLIP} (TAO-FineTuned) & 60.2 & 12.6 & 27.5\\
        \texttt{ViFi-CLIP} (ActivityNet-FT) & 32.7 & 58.2 & 30.4 \\
        \bottomrule
    \end{tabular}
    \vspace{-3em}
\end{table}

\section{Comparison with Single Label Open Vocabulary Baselines}
\label{sec:comparisonopen}

\noindent STAN~\cite{liu2023revisiting} is intended for fully fine-tuned setting; it doesn't report any zero-shot results.  Open V-CLIP~\cite{weng2023open} is trained to solve single label classification. In contrast, our goal is to perform multi-label classification in the zero-shot setting. For comparison we provide zero-shot results for both. As STAN doesn't provide pretrained weights, we train it on our training dataset. For Open V-CLIP, we use author provided pretrained weights.

\begin{table}
    
    \caption{\textbf{Open Vocabulary baselines} (Peak F1)} \label{tab:ovbaselines}
    \scriptsize
    \centering
    \renewcommand{\arraystretch}{0.9}
    \setlength{\tabcolsep}{2pt}
    \hspace{-3.4em}
    \begin{tabular}{l|cc}
        \toprule
        \textbf{Method} & \textbf{TAO} & \textbf{ActivityNet} \\
        \midrule
        \texttt{STAN} (K400+YT8M, ours) & 58.1 & 27.6 \\
        \texttt{Open V-CLIP} (original) & 43.9 & 50.2 \\
        \cmidrule{1-3}
        {\bf Ours} & {\bf 59.6} & \textbf{ 52.6} \\
        \bottomrule
    \end{tabular}
    \vspace{-1.5em}
\end{table}

\section{Comparison with Multi-Modal LLMs}
\label{sec:comparisonllava}

Recently in the literature~\cite{yousaf2024videoprompter, llovi}, general purpose multi-modal LLMs have been demonstrated to achieve competitive performance across a range of video understanding tasks. They are not practical for our setting, since they impose a significant computation cost, however to demonstrate the advantage of our solution over multi-modal LLMs,
we construct two LLaVA-based inference baselines. For the first, we prompt LLaVA regarding the presence of a class label in video frames. For second, we closely follow our synthetic label generation pipeline (see Section~\ref{sec:pipelinesupp}) and generate frame captions using LLaVA. The captions are then classified using CLIP's text encoder. The results in Table~\ref{tab:llava} show that LLaVA performs significantly worse than our method, even when no synthetic labels are used for training.

Both LLaVA based approaches require running the Multi-Modal LLM for every video at inference. Additionally, for the first approach, we need to run it for every label in the validation vocabulary. In contrast, for our method the LLM is not used during inference, but only when a new label is added to the classification vocabulary.

\textcolor{red}{}

\begin{table}[h]
    \caption{\textbf{LLaVA Inference baselines} \footnotesize{(Peak F1-Score)}} \label{tab:llava}
    \vspace{-0.5em}
    \scriptsize
    \centering
    \begin{tabular}[t]{l|cc|c}
        \toprule
        \textbf{Method} & TAO & ActivityNet & Inference Time (1$\times$ A100)\\
        \midrule
        LLaVA Classification (yes/no polling every class) & 27.8 & 11.5 & 10 min+\\
        LLaVA Captioning + CLIP Classification & 47.2 & 34.7 & 10s \\
        \cmidrule{1-4}
        {\bf Ours} & {\bf 59.6} & 52.6 & 0.25s\\
        {\bf Ours w/ added synthetic labels during training} & 56.6 & {\bf 53.8} & 0.25s \\
        \bottomrule
    \end{tabular}
    \vspace{-2em}
\end{table}

\section{Zero-Shot Single Label Action Classification}

Our open vocabulary model though trained for multi-label classification is also competitive 
 (see row \textbf{(a)} in Table~\ref{tab:singlelab})  on the 
 zero-shot single label action classification task. 

\noindent A key difference between our approach and prior single label classification works tailored to this problem is our use of binary classification losses, which is essential for multi-label classification but is not optimal for single label classification, which only requires ranking the labels. In order to match the setting of prior works, we also train our model on only Kinetics-400 using Cross-Entropy loss and provide results in row \textbf{(b)} to show that it can exceed prior work such as ViFi-CLIP~\cite{hanoonavificlip}. 

\label{sec:singlelabel}
\begin{table}[h]
\vspace{-1em}
\centering
\footnotesize
\begin{tabular}{lrrr} 
\toprule
 \textbf{Model} & \multicolumn{1}{l}{\textbf{UCF101}} & \multicolumn{1}{l}{\textbf{HMDB51}} & \multicolumn{1}{l}{\textbf{Kinetics600}} \\ 
\midrule
\multicolumn{4}{l}{\textit{No video data used in training}} \\
CLIP & 61.7 & 37.5 & 63.5 \\
CLIP + LLM & 73.8 & 46.1 & 64.8\\ 
\midrule
\multicolumn{4}{l}{Trained on \textit{YouTube8M + Kinetics400}} \\
{\bf (a) Ours}& 74.1 & 53.2 & 67.7\\
\midrule
\multicolumn{4}{l}{Trained on \textit{Kinetics400}} \\
Vi-Fi CLIP~$^{\ast}$ & 77.5 & 51.8 & 71.2 \\
{\bf (b) Ours (Cross-Entropy Loss)} & \textbf{79.0} & \textbf{54.5} & \textbf{72.8} \\
\bottomrule
\end{tabular}
\vspace{0.2em}
\caption{Results on single label action classification datasets. Top-1 Accuracy is reported for all datasets. $^{\ast}$ Results reported in ViFi CLIP~\cite{hanoonavificlip}}
\label{tab:singlelab}
\vspace{-2em}
\end{table}

\section{Zero-Shot Evaluation on EgoCentric tasks}
\label{sec:egocentric}

We provide results for scenario classification on Ego4d and verb \& noun identification for Epic-Kitchens (unseen kitchens).

    \begin{table}[h]
    \vspace{-1.9em}
    \caption{\textbf{Egocentric} \footnotesize{(Peak F1)}} \label{tab:egocentric}
    \vspace{-0.5em}
    \scriptsize
    \centering
    \setlength{\tabcolsep}{3pt}
    \renewcommand{\arraystretch}{0.8}
    \begin{tabular}[t]{l|ccc}
        \toprule
        \textbf{Method} & Ego4D & EK-unseen (Verbs) & EK-unseen (Nouns) \\
        \midrule
        CLIP     & 45.3 & 16.5 & 32.8 \\
        CLIP+LLM & 48.7 & 20.3 & 39.1 \\
        \cmidrule{1-4}
        {\bf Ours} & {\bf 51.9} & \textbf{22.1} & \textbf{40.5} \\
        \bottomrule
    \end{tabular}
    \vspace{-1.5em}
\end{table}

\section{Additional Ablations for Label Encoder}
\label{sec:labablateextra}

We conducted additional ablations for the LLM adapter and LLM-VLM connector, with results shown in Table~\ref{tab:textablate}. Due to time constraints, we trained for only 10k steps, nearing convergence. We find that LoRA saturates after rank 2 and performs worse than prompt learning for Zero-Shot Generalization. Our prompting transformer outperforms MLP \& Linear connectors.

\begin{table}[t]
    \caption{\textbf{LLM Adaptation Ablations} \footnotesize{(Peak F1-Score, 10k training steps)}} \label{tab:textablate}
    \vspace{-0.5em}
    \scriptsize
    \centering
    \renewcommand{\arraystretch}{0.83}
    \begin{tabular}[t]{ccc|cc}
        \toprule
        Steps & LLM Adapter & LLM-VLM Connector & TAO & ActivityNet \\
        \midrule
        10k & LoRA (r=2) & Prompting Transformer & 57.9 & 49.4\\
        10k & LoRA (r=4) & Prompting Transformer & 58.0 & 46.9 \\
        10k & LoRA (r=2) & Linear & 46.2 & 38.9 \\
             \midrule
         10k & Prompts & Linear & 48.4 & 35.2 \\
         10k & Prompts & MLP & 52.9 & 44.7 \\
        10k  & \textbf{Prompts} & \textbf{Prompting Transformer} &  {\bf 58.3} & \textbf{50.8} \\
        \textcolor{gray}{50k} & \textcolor{gray}{Prompts} & \textcolor{gray}{Prompting Transformer} &   \textcolor{gray}{59.6} & \textcolor{gray}{52.6} \\
        \bottomrule
    \end{tabular}
    \vspace{-2em}
\end{table}

\section{Additional Ablations for Temporal Encoder}
\label{sec:tempablateextra}

We designed an alternative \noindent version of our architecture with serial blocks instead of parallel and train the model again. The results (Table~\ref{tab:tempablate}) indicate that parallel blocks outperform serial blocks. As our main goal is open vocabulary classification, our temporal ablations (Table 4) are focused on regularization and related aspects. Exhaustive temporal architecture ablations are beyond the scope of a single paper, and different aspects of temporal modeling have been studied previously (\cite{vindlu}, \cite{liu2023mug}, \cite{vidla}).

\begin{table}
    \vspace{-2em}
    \caption{\textbf{Temporal} \footnotesize{(Peak F1)}} \label{tab:tempablate}
    \vspace{-0.5em}
    \centering
    \begin{tabular}[t]{c|cc}
        \toprule
        Temporal Adapter & TAO & ActivityNet \\
        \midrule
        Serial (n=4) & 53.2 & 41.5 \\
        \textbf{Parallel} (n=4) &  {\bf 59.6} & \textbf{52.6} \\
        \bottomrule
    \end{tabular}
    \vspace{-3em}
\end{table}

\section{Comparison of Computational Costs}
\label{sec:computecost}

\begin{table}[h]
\vspace{-2em}
    \caption{\textbf{Computational Costs}} \label{tab:compute}
    \vspace{-1em}
    \scriptsize
    \centering
    \begin{tabular}[t]{c|cc|c}
        \toprule
        
        \textbf{Method} & \multicolumn{2}{c}{\textbf{Training (YT8M+K400)}} & \textbf{Inference time} \\ 
        & Time & Mem/GPU & (batch size = 32)\\
        \midrule
        \textbf{ViFi-CLIP} & 36 Hrs & 11.0 GB & 338ms\\
        {\bf Ours} & 40 Hrs & 16.5 GB& 393ms\\
        \bottomrule
    \end{tabular}
    \vspace{-1em}
\end{table}

\noindent Training time (on 16 $\times$ A100 GPUs) on YT-8M + K400 for our method is about 10\% higher  than ViFi-CLIP baseline. Inference on 1 RTX8000 is about 16\% slower (batch size=$32$, using \texttt{torchinfo}
 package). Text embeddings for class labels can be pre-computed, only video features need computing on the fly during inference.

\section{Synthetic Label Generation Pipeline}
\label{sec:pipelinesupp}

\begin{figure*}[h]
    \centering
    \begin{subfigure}[b]{\linewidth}
        \includegraphics[width=\linewidth,clip]{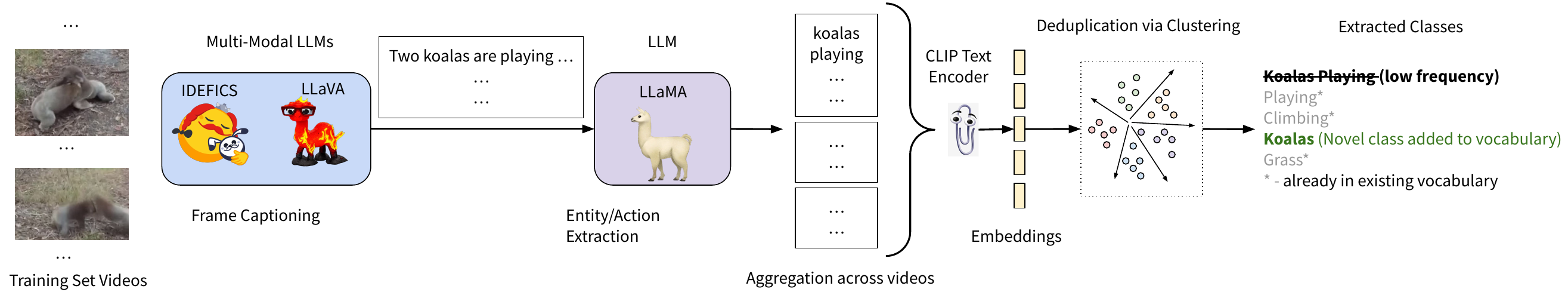}
        \caption{Expanding the label vocabulary.}
        \label{fig:vocabexpansion}
    \end{subfigure}
    \vspace{1em} 
    \begin{subfigure}[b]{\linewidth}
        \includegraphics[width=\linewidth,clip]{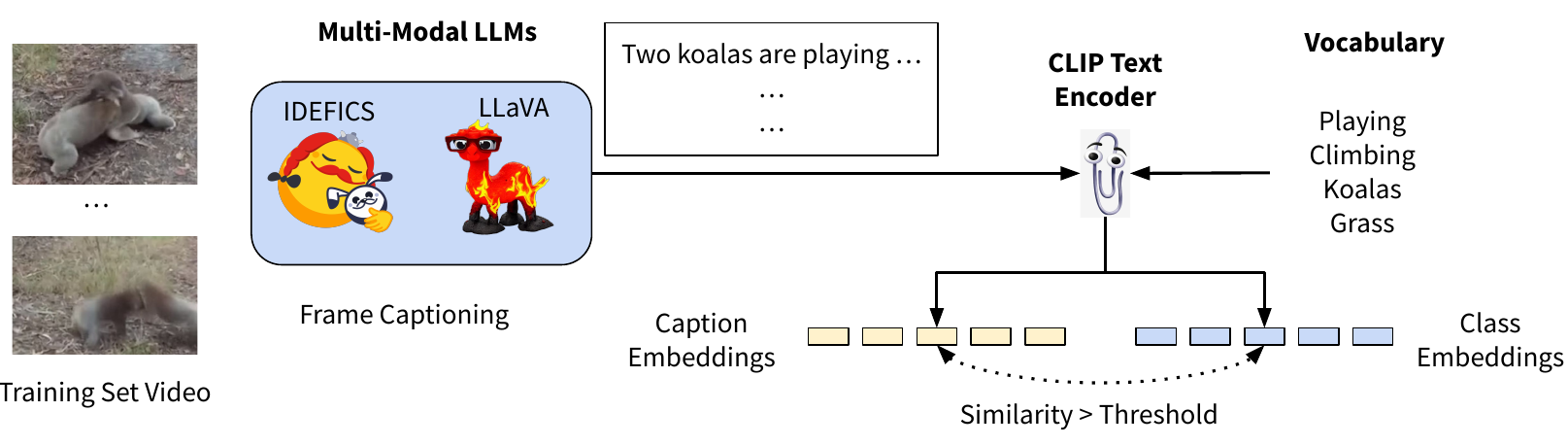} %
        \caption{Assigning labels to videos.} %
        \label{fig:labelassignment}
    \end{subfigure}
    \caption{Incorporating synthetic labels into our training sets enhances our open vocabulary performance further. \textbf{(a) Vocabulary Expansion:} We have developed a pipeline to automatically extract action and object labels from a vast video dataset utilizing foundation models. For captioning video frames, we employ Multi-Modal Large Language Models (LLMs), specifically IDEFICS and LLaVA. Subsequently, LLaMA is prompted to distill object and action class labels from these captions. We aggregate these labels across videos and remove duplicates through clustering, forming a classification vocabulary. \textbf{(b) Label Assignment:} Labels from the vocabulary are aligned with the generated video captions using the text encoder from CLIP.} %
    \label{fig:labelpipeline}
\end{figure*}
As illustrated in Figure~\ref{fig:labelpipeline}  our synthetic labeling pipeline consists of four steps: caption generation, concept extraction, vocabulary expansion and label assignment. The caption generation process employs off-the-shelf multi-modal LLMs and is straightforward. 

For the second step, we prompt LLaMA 2-13B-Chat to extract concept labels for videos from captions generated in Stage 1.The LLM prompt for extracting labels from captions is provided in Listing~\ref{lst:label}. We provide 3 in-context examples and the LLM is prompted to extract concept labels from the fourth video's captions. 

As LLMs identify a large number of concepts, including many near-duplicates, a cleanup step is necessary to minimize these issues. We utilize the CLIP text encoder to obtain embeddings for all the identified concepts across the dataset. K-Means clustering is applied to the embeddings to cluster them into groups. For each group, we replace the labels with the most frequently observed concepts from that group. This works reasonably well, and CLIP text encoder is excellent at detecting near-duplicate visual concepts. A random sample of identified clusters are shown in Table~\ref{tab:clusters}. 

Finally, we reuse the CLIP text encoder to match labels from the deduplicated vocabulary back to videos, which are represented by their captions. In order to reduce domain shift between captions and the labels, standard CLIP prompt template \texttt{"a video of \{label\}"} is used.

For extracting extra action labels, we use IDEFICS-9b-Instruct model~\cite{laurençon2023obelics} and LLaVA~\cite{liu2023llava} to caption the videos. Both models are based on LLaMA LLMs, with IDEFICS trained using the interleaved image text dataset OBELICS, while LLaVA is trained on a mix of image-caption data and instruction following data created using GPT-3 and image annotations. This stage is followed by LLaMA 2-13b-chat~\cite{touvron2023llama} to extract the labels from the captions. OpenAI CLIP B/32 is used to clean up label assignment to videos.

\begin{table}[h]
\centering
\scriptsize
\begin{tabular}{|p{0.15\linewidth} | p{0.8\linewidth}|} 
\toprule
\textbf{Airlines} & `american airlines', `delta air lines', `southwest airlines', `singapore airlines', `air france', `emirates (airline)', `british airways', `carnival cruise line' \\
\midrule
\textbf{Grilling} & `barbequing', `cooking on campfire', `grilling', `barbecue' \\ 
\midrule
\textbf{Lego} & `legoland', `lego star wars', `lego minecraft', `lego duplo', `the lego group', `lego friends', `lego batman: the videogame', `lego', `lego batman 3: beyond gotham', `lego ninjago', `lego minifigure', `playmobil', `lego marvel super heroes', `lego city', `lego legends of chima' \\ 
\midrule
\textbf{Playground} & `playing on a playground', `playground', `amusement park', `amusement arcade', `amusement ride', `water park', `ferris wheel' \\ 
\midrule
\textbf{Video Games} & `gears of war (video game)', `jill valentine', `gears of war', `silent hill 2', `resident evil 2', `hitman: absolution', `gears of war 2', `resident evil 5', `resident evil 3: nemesis', `resident evil', `resident evil (1996 video game)', `resident evil (2002 video game)' \\ 
\midrule
\textbf{Water Slide} & `water sliding', `riding water slide', `water slide' \\ 
\bottomrule
\end{tabular}
\vspace{0.2em}
\caption{Sampled clusters among concepts identified by the captioning + LLM steps of the label generation pipeline. CLIP Text encoder features were used to cluster the concepts for de-duplication. Cluster names assigned in the left column are only used for illustration.}
\label{tab:clusters}
\end{table}

\clearpage

\begin{myverbatim}{Our LLM prompt for extracting action labels from video captions}
Following is the description of a video. Output a numbered list of verbs 
representing visual actions performed in the video. Do not add any explanation.

Video 1 description:
1. A group of people riding motorcycles at night.
2. A motorcycle is lit up with blue lights.
3. A person is riding a bike at night.
4. A motorcycle parked on the street at night.
5. A group of people are gathered in a dimly lit room.
6. A motorcycle parked in a dark room.
7. A motorcycle is parked in a dark room.
8. A person is riding a bike at night.

Verbs Found:
1. riding motorcycle
2. riding bike

Following is the description of a video. Output a numbered list of verbs 
representing visual actions performed in the video. Do not add any explanation.

Video 2 description:
1. A man is performing on stage with a band.
2. A group of men are performing on a stage.
3. A man with a microphone is performing on stage.
4. A group of young men performing on stage.
5. A man is singing on a stage with a band.
6. A man is playing a guitar on a stage.
7. A man and a woman are performing on stage.
8. A dark room with a bright light shining on it.

Verbs Found:
1. performing on stage
2. singing on stage
3. playing guitar

Following is the description of a video. Output a numbered list of verbs 
representing visual actions performed in the video. Do not add any explanation.


Video 3 description:
1. A person is putting lotion on another person's hand.
2. A person is putting nail polish on another person's nails.
3. A person is putting nail polish on their nails.
4. A person is holding a ball point pen.
5. A person is writing on a piece of paper.
6. A person is holding another person's hand.
7. A person is putting a ring on another person's finger.
8. A black screen with a white frame.

Verbs Found:
1. putting lotion
2. putting nail polish
3. writing
4. putting ring

Following is the description of a video. Output a numbered list of verbs 
representing visual actions performed in the video. Do not add any explanation.


Video 4 description:
<output_captions>

Verbs Found:
\end{myverbatim}

\section{Baselines}
\label{sec:baselinessupp}

\subsection{CLIP + LLM Frozen Baseline}
This baseline is an extension of and inspired by prior works~\cite{menon2023visual, pratt2023does} that utilize LLMs to prompt CLIP for image classification to our problem of video classification. The LLM is prompted to generate class descriptors utilizing its extensive world knowledge. CLIP can then be used to match these descriptors to video frames to classify them.

\begin{figure*}[t]
\centering
\begin{subfigure}{0.6\linewidth}
    \includegraphics[width=\linewidth,trim={6em 6em 10em 3em},clip]{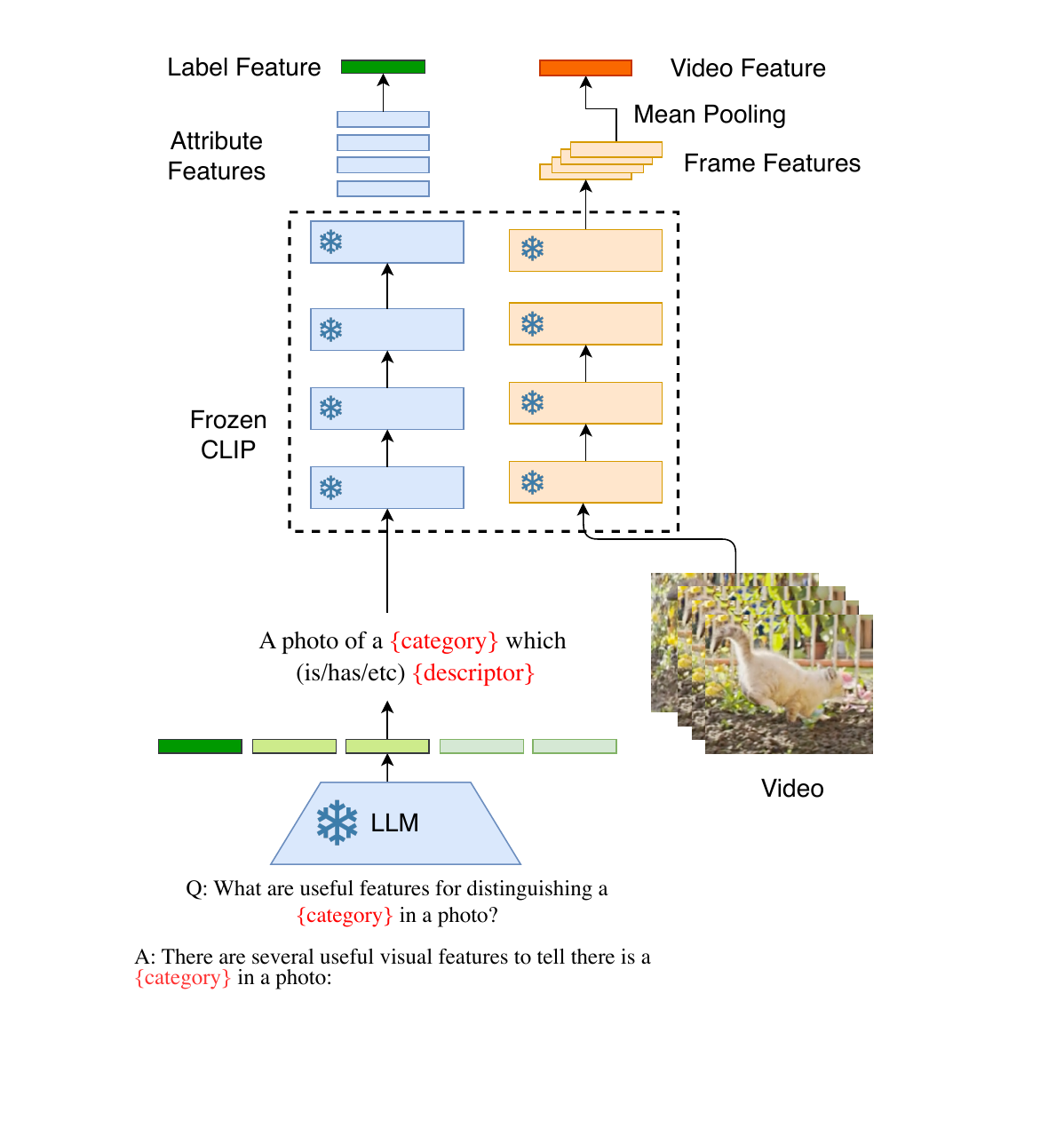}
    \caption{CLIP + LLM Baseline Architecture.}
    \label{fig:baselinearch}
\end{subfigure}
\hfill
\begin{subfigure}{0.38\linewidth}
    \includegraphics[width=\linewidth,trim={0em 0em 0em 0em},clip]{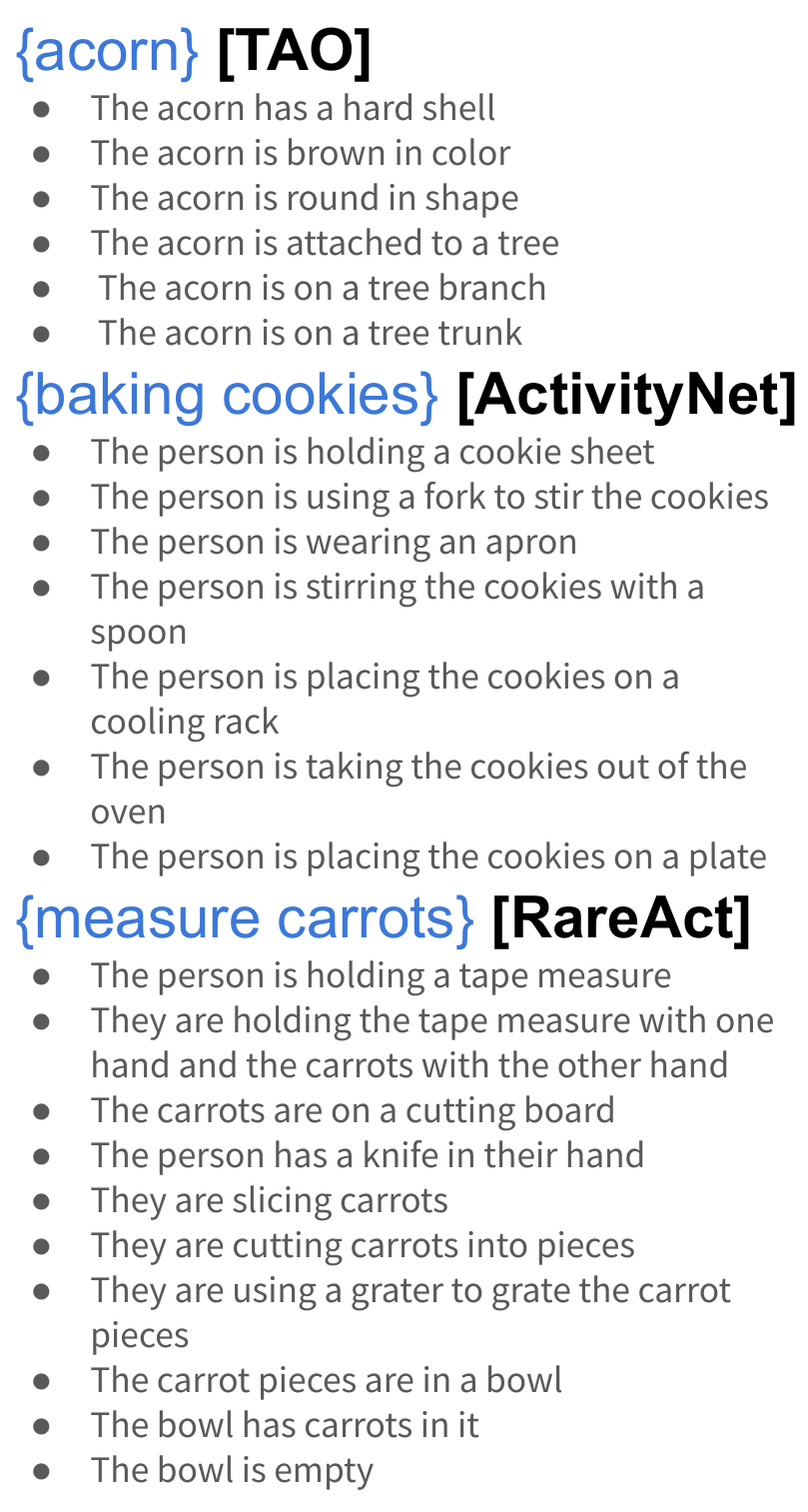}
    \vspace{-1.75em}
    \caption{Sample LLM generated Class Descriptors.}
    \label{fig:baseoutputs}
\end{subfigure}
\\
\begin{subfigure}{0.7\linewidth}
    \includegraphics[width=\linewidth,trim={0em 0em 0 0em},clip]{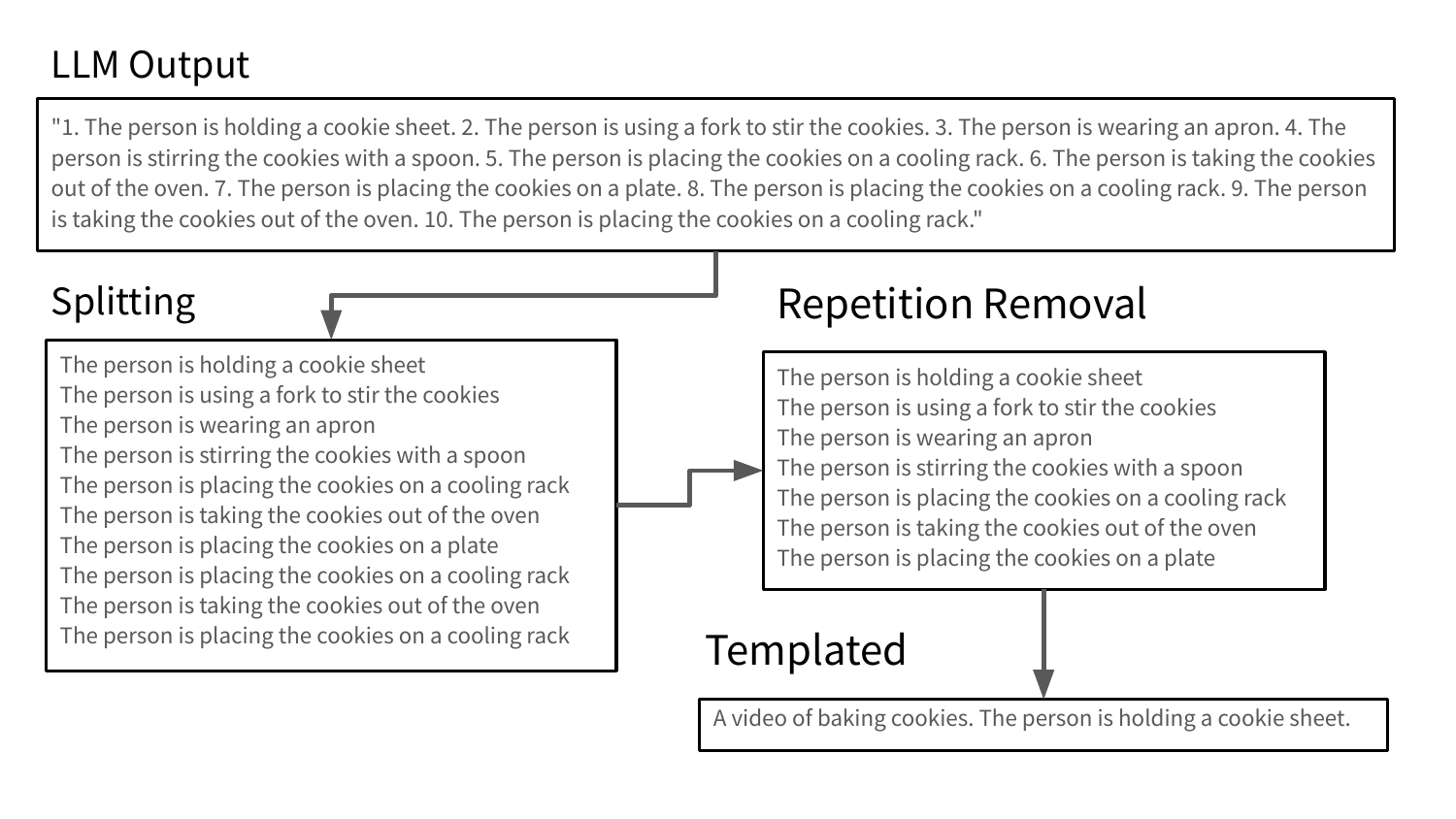}
    \vspace{-1.75em}
    \caption{Process for cleaning up LLM output to generate CLIP Prompts.}
    \label{fig:basepostproc}
\end{subfigure}
\vspace{-0.8em}
\caption{ Here we illustrate the CLIP+LLM baseline discussed in the main paper. \textbf{(a)} Architecture consist of frozen CLIP and LLM model. The LLM is prompted to generate class descriptors to assisst CLIP.  \textbf{(b)} Some sample class descriptors generated by the LLM. \textbf{(c)} Process for converting the raw LLM output text to attribute prompts for CLIP. Firstly, the raw text is split into separate list items, then repetitions (which LLMs are prone to) are removed and finally standard CLIP prompting templates are used to combine the class name with the descriptor.} 
\vspace{-1em}
\label{fig:baseline}
\end{figure*}

The overall process is illustrated in Figure~\ref{fig:baselinearch}. Unlike the image classification setting it includes a mean pooling operation across frames to get the video level feature. Note that the LLM prompt is designed to elicit output in the form of a list. This simplifies the post-processing of the text output to generate CLIP prompts (Figure~\ref{fig:basepostproc}). Firstly, the text is split into each item of the list, followed by removal of repetitions (common for this generation of LLMs). Finally we use a standard CLIP prompting template to incorporate both the class label and the descriptor. Sample descriptors for some classes from our downstream datasets are provided in Figure~\ref{fig:baseoutputs}. CLIP + LLM is a reasonably and consistently strong baseline across datasets, as it inherits CLIP's robustness.

A key limitation of this baseline is the frequency of LLM failures. Different classes of failures such as getting trapped in a repetition loop, generating descriptors which are not visual, and semantic confusion are common.  An end-to-end trainable approach could potentially alleviate some of these issues.

\subsection{CoOp: Context Optimization}

CoOp~\cite{zhou2022learning} learns prompts for CLIP's text encoder to adapt it for image classification. This is a parameter efficient adaptation method since it has very few learnable parameters. We extend it to the video setting by utilizing mean pooling across frame in the vision encoder and learnable prompts in the text encoder. (See Figure~\ref{fig:coop})

\subsection{DualCoOp}
\begin{figure*}[h]
\centering
\begin{subfigure}{0.48\linewidth}
    \includegraphics[width=\linewidth]{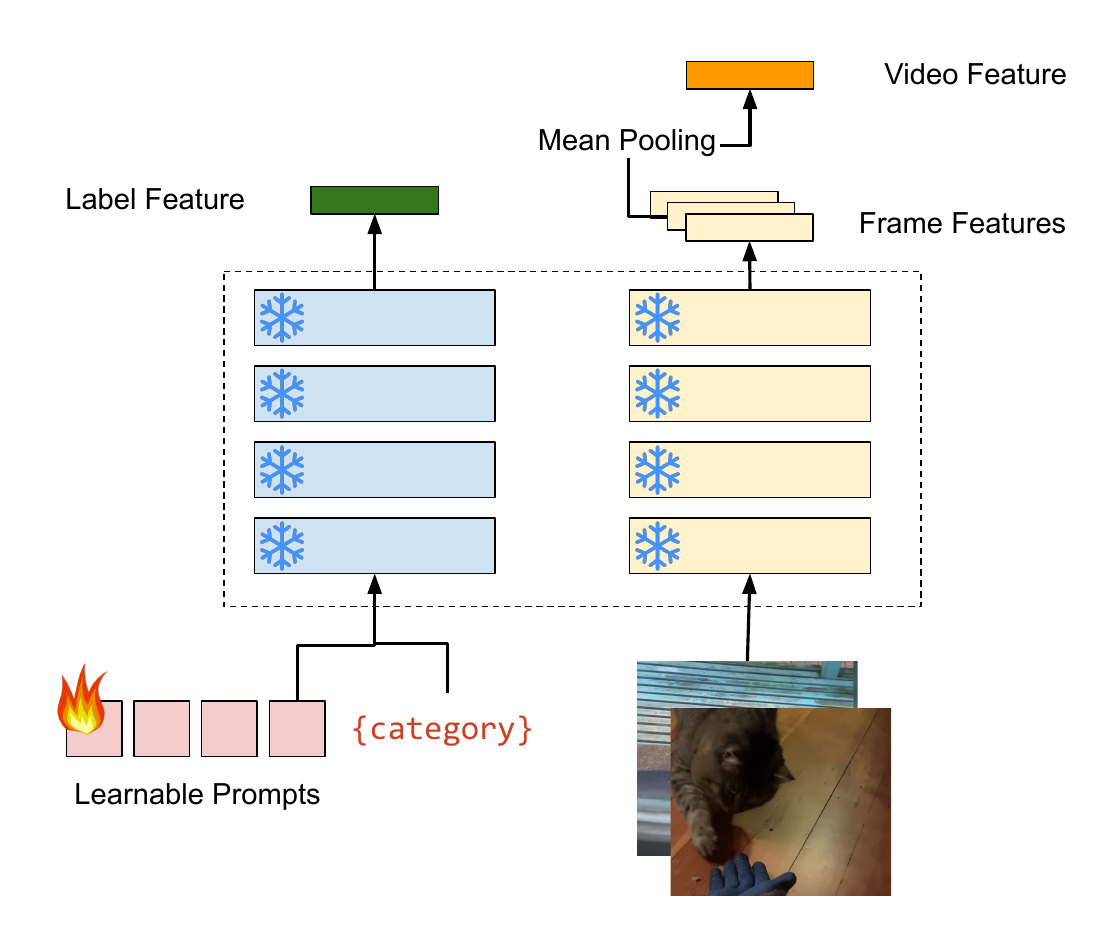}
    \caption{CoOp.}
    \label{fig:coop}
\end{subfigure}
\hfill
\begin{subfigure}{0.48\linewidth}
    \includegraphics[width=\linewidth]{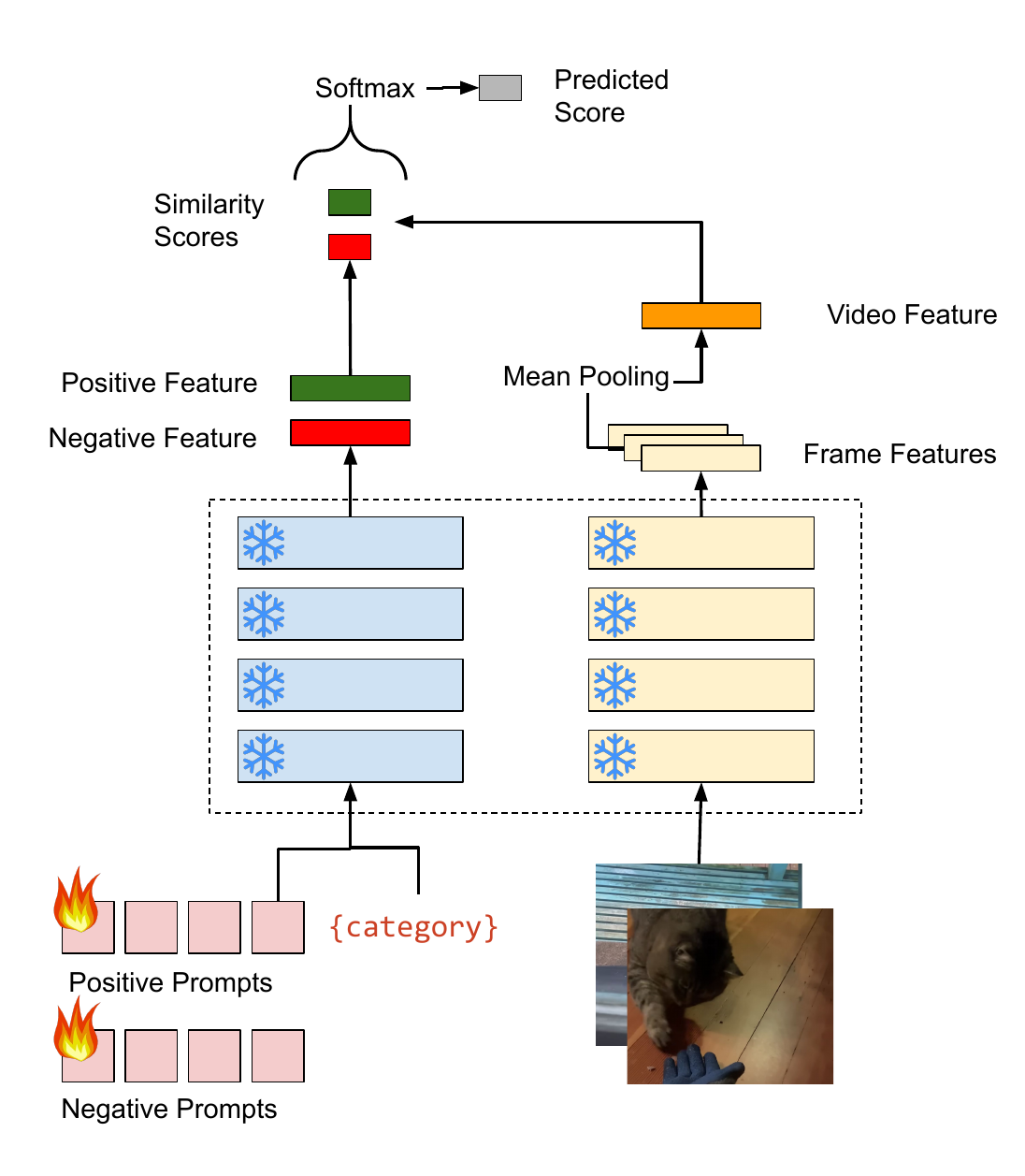}
    \vspace{-1.75em}
    \caption{Dual CoOp.}
    \label{fig:dualcoop}
\end{subfigure}
\vspace{-0.8em}
\caption{Trainable CLIP based baselines without an LLM \textbf{(a) CoOp} utilizes learnable prompts on the text encoder side to guide the model towards classification task.   \textbf{(b) Dual CoOp} is designed for multi-label classification and utilizes learnable positive and negative prompts to generate a probability score for each label.} 
\vspace{-1em}
\label{fig:coops}
\end{figure*}

DualCoOp~\cite{sun2022dualcoop} refines prompt learning for the multi-label setting, with both positive and negative learnable prompts. A label is matched to a video if the similarity score of the video features with the features for the positive prompts is higher than for the negative prompts. A soft prediction score can be obtained by taking a softmax across the two similarity values. (See Figure~\ref{fig:dualcoop})

\clearpage

\section{Implementation Details}
\label{sec:implementsupp}

We use Open AI CLIP-B/32~\cite{radford2021learning} as the backbone VLM and Flan-T5-XL~\cite{chung2022scaling} as the promptable LLM. We sample 8 frames per clip during training, and during evaluation we use 4 clips per video (total of 32 frames). Our best model uses 4 learnable LLM Prompts, and 4 layers of temporal modeling. (Following the notation from the method section, $N=4$, $K=5$, $L=5$ and $T=4$). Following the Flan-T5 text generation instructions we use the following prompt:

\noindent \texttt{Q: What are useful features for distinguishing a \textcolor{red}{\{\textbf{label}\}} in a photo?}

\noindent \texttt{A: There are several useful visual features to tell about a \textcolor{red}{\{\textbf{label}\}} in a photo:}

\noindent \texttt{1. \textcolor{blue}{\textbf{<extra\_id\_0>}}}

\noindent Where \textcolor{red}{\{\textbf{label}\}} represents a given class label and \texttt{\textcolor{blue}{\textbf{<extra\_id\_0>}}} is T5 decoder's start token ID. In case of the frozen baseline, when presented with this prompt, the LLM produces text that consists of a list of label attributes, which can be parsed and separated into different prompts for CLIP. We mimic this approach in our learnable version by chunking LLM output into $K$ groups of $L$ tokens each.

We use a batch size of 12 videos/GPU across 32 A100 GPUs (Total Batch Size=384) and train all models for $30000$ training steps and evaluate at $10000$, $20000$ and $30000$ steps, providing the best results for each methods across these 3 checkpoints. We do this to ensure a fair comparison as our different baselines and methods have widely varying number of trainable parameters, it would not be a fair comparison to use the same number of steps for each.

\subsection{Datasets Used}
\vspace{-0.5em}

For training, we use YouTube-8M and Kinetics datasets. For evaluation we use TAO (Tracking Any Object) dataset for Object Classification and ActivityNet for action classification. We also leverage the RareAct dataset in a novel way by using their noun and verb labels to generate 3 labels for each clip, noun, verb and noun-verb combination. For YouTube-8M, we use the human verified validation set for reporting results. Overall these evaluation datasets cover a wide range of entities and actions, providing a comprehensive evaluation of open vocabulary multi-label video classification capabilities.

\begin{table}[h]
\centering
\vspace{-2em}
  \renewcommand{\arraystretch}{0.9}
\begin{tabular}{lrrc} 
\toprule
\textbf{Dataset} & \multicolumn{1}{l}{\# \textbf{Videos}} & \multicolumn{1}{l}{\# \textbf{Classes}} & \multicolumn{1}{l}{\# \textbf{Labels/Video}} \\ 
\midrule
\textit{Training Datasets} \\
YouTube-8M & 2,285,432 & 2429 & 2.9 \\ 
 + Generated Labels & 2,285,432 & 3281 & 6.7 \\ 
Kinetics 400 & 246,245 & 400 & 1 \\ 
 + Generated Labels & 246,245 & 1355 & 4.5 \\ 
\midrule
\textit{Test Datasets} \\
YT-8M Segments Val & 42,407 & 1000 & 1.05 \\ 
TAO & 655 & 1230 & 1.44 \\ 
ActivityNet & 4,593 & 200 & 1.01 \\ 
RareAct & 905 & 214 & 3.02 \\
\bottomrule
\end{tabular}
\label{tab:datasets}
\caption{Details about datasets used for training. For YouTube-8M and Kinetics, we also generate additional labels for training using our synthetic labelling pipeline.}
\vspace{-4em}
\end{table}

\subsection{Training Details}

We use Open AI CLIP-B/32 as the Vision-Language model and Google Flan-T5 XL as the promptable LLM. 

We use \texttt{AdamW} optimizer for training with a base learning rate of $0.00001$. Weight decay for newly initialized layers is set to  $0.0000001$ and $0.0$ for CLIP initialized layers. Weight regularization loss weight for STAN's spatial attention layers is set to $\lambda = 0.000001$. Cosine decay learning rate scheduler with warmup is used. Total training length is $30,000$ steps including $2,000$ steps of warmup.

\section{Qualitative Results}
\label{sec:qualitative}
\vspace{-2em}
\begin{figure}[h]
    \centering
    \includegraphics[width=0.4\linewidth]{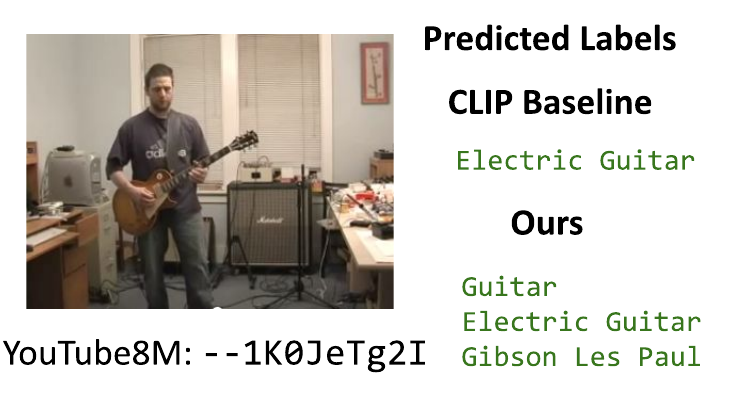}
    \includegraphics[width=0.4\linewidth]{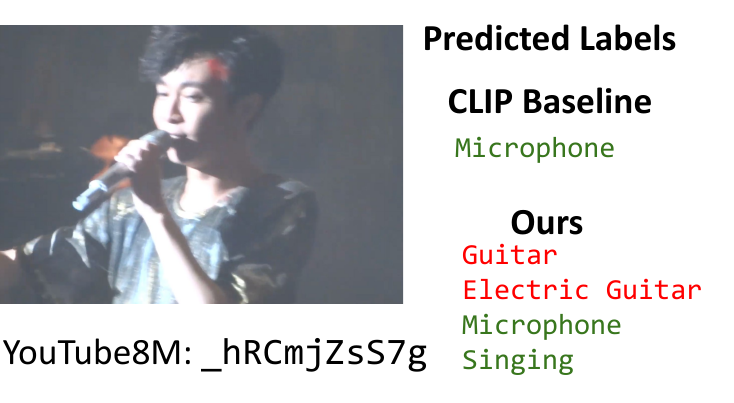}
    \label{fig:qualitative}
\end{figure}

\bibliographystyle{splncs04}
\bibliography{main}

\begin{thebibliography}{10}
\providecommand{\url}[1]{\texttt{#1}}
\providecommand{\urlprefix}{URL }
\providecommand{\doi}[1]{https://doi.org/#1}

\bibitem{abuelhaija2016youtube8m}
Abu-El-Haija, S., Kothari, N., Lee, J., Natsev, P., Toderici, G., Varadarajan, B., Vijayanarasimhan, S.: Youtube-8m: A large-scale video classification benchmark (2016)

\bibitem{gberta_2021_ICML}
Bertasius, G., Wang, H., Torresani, L.: Is space-time attention all you need for video understanding? In: Proceedings of the International Conference on Machine Learning (ICML) (July 2021)

\bibitem{vindlu}
Cheng, et~al.: {VindLU}: A recipe for effective video-and-language pretraining. In: CVPR (2023)

\bibitem{chung2022scaling}
Chung, H.W., Hou, L., Longpre, S., Zoph, B., Tay, Y., Fedus, W., Li, Y., Wang, X., Dehghani, M., Brahma, S., Webson, A., Gu, S.S., Dai, Z., Suzgun, M., Chen, X., Chowdhery, A., Castro-Ros, A., Pellat, M., Robinson, K., Valter, D., Narang, S., Mishra, G., Yu, A., Zhao, V., Huang, Y., Dai, A., Yu, H., Petrov, S., Chi, E.H., Dean, J., Devlin, J., Roberts, A., Zhou, D., Le, Q.V., Wei, J.: Scaling instruction-finetuned language models (2022)

\bibitem{dave2020tao}
Dave, A., Khurana, T., Tokmakov, P., Schmid, C., Ramanan, D.: Tao: A large-scale benchmark for tracking any object. In: Computer Vision--ECCV 2020: 16th European Conference, Glasgow, UK, August 23--28, 2020, Proceedings, Part V 16. pp. 436--454. Springer (2020)

\bibitem{desai2021redcaps}
Desai, K., Kaul, G., Aysola, Z., Johnson, J.: Redcaps: Web-curated image-text data created by the people, for the people. arXiv preprint arXiv:2111.11431  (2021)

\bibitem{fan2023improving}
Fan, L., Krishnan, D., Isola, P., Katabi, D., Tian, Y.: Improving clip training with language rewrites. In: NeurIPS (2023)

\bibitem{fang2021clip2video}
Fang, H., Xiong, P., Xu, L., Chen, Y.: Clip2video: Mastering video-text retrieval via image clip. arXiv preprint arXiv:2106.11097  (2021)

\bibitem{gorti2022x}
Gorti, S.K., Vouitsis, N., Ma, J., Golestan, K., Volkovs, M., Garg, A., Yu, G.: X-pool: Cross-modal language-video attention for text-video retrieval. In: Proceedings of the IEEE/CVF conference on computer vision and pattern recognition. pp. 5006--5015 (2022)

\bibitem{Gupta_2023_CVPR}
Gupta, R., Roy, A., Christensen, C., Kim, S., Gerard, S., Cincebeaux, M., Divakaran, A., Grindal, T., Shah, M.: Class prototypes based contrastive learning for classifying multi-label and fine-grained educational videos. In: Proceedings of the IEEE/CVF Conference on Computer Vision and Pattern Recognition (CVPR). pp. 19923--19933 (June 2023)

\bibitem{he2022asm}
He, B., Yang, X., Kang, L., Cheng, Z., Zhou, X., Shrivastava, A.: Asm-loc: Action-aware segment modeling for weakly-supervised temporal action localization. In: Proceedings of the IEEE/CVF conference on computer vision and pattern recognition. pp. 13925--13935 (2022)

\bibitem{activitynet}
Heilbron, F.C., Niebles, J.C.: Collecting and annotating human activities in web videos. In: Proceedings of International Conference on Multimedia Retrieval. p. 377–384. ICMR '14, Association for Computing Machinery, New York, NY, USA (2014). \doi{10.1145/2578726.2578775}, \url{https://doi.org/10.1145/2578726.2578775}

\bibitem{hu2022lora}
Hu, E.J., yelong shen, Wallis, P., Allen-Zhu, Z., Li, Y., Wang, S., Wang, L., Chen, W.: Lo{RA}: Low-rank adaptation of large language models. In: International Conference on Learning Representations (2022), \url{https://openreview.net/forum?id=nZeVKeeFYf9}

\bibitem{jia21b}
Jia, C., Yang, Y., Xia, Y., Chen, Y.T., Parekh, Z., Pham, H., Le, Q., Sung, Y.H., Li, Z., Duerig, T.: Scaling up visual and vision-language representation learning with noisy text supervision. In: Meila, M., Zhang, T. (eds.) Proceedings of the 38th International Conference on Machine Learning. Proceedings of Machine Learning Research, vol.~139, pp. 4904--4916. PMLR (18--24 Jul 2021), \url{https://proceedings.mlr.press/v139/jia21b.html}

\bibitem{jia2021scaling}
Jia, C., Yang, Y., Xia, Y., Chen, Y.T., Parekh, Z., Pham, H., Le, Q., Sung, Y.H., Li, Z., Duerig, T.: Scaling up visual and vision-language representation learning with noisy text supervision. In: International conference on machine learning. pp. 4904--4916. PMLR (2021)

\bibitem{Kaul23}
Kaul, P., Xie, W., Zisserman, A.: Multi-modal classifiers for open-vocabulary object detection. In: International Conference on Machine Learning (2023)

\bibitem{kay2017kinetics}
Kay, W., Carreira, J., Simonyan, K., Zhang, B., Hillier, C., Vijayanarasimhan, S., Viola, F., Green, T., Back, T., Natsev, P., Suleyman, M., Zisserman, A.: The kinetics human action video dataset (2017)

\bibitem{Swetha_safellava}
Kim, Y., Swetha, S., Kagdi, F., Shah, M.: Safe-llava: A privacy-preserving vision-language dataset and benchmark for biometric safety. arXiv preprint arXiv:2509.00192  (2025)

\bibitem{kojima2022large}
Kojima, T., Gu, S.S., Reid, M., Matsuo, Y., Iwasawa, Y.: Large language models are zero-shot reasoners. In: Oh, A.H., Agarwal, A., Belgrave, D., Cho, K. (eds.) Advances in Neural Information Processing Systems (2022), \url{https://openreview.net/forum?id=e2TBb5y0yFf}

\bibitem{laurençon2023obelics}
Laurençon, H., Saulnier, L., Tronchon, L., Bekman, S., Singh, A., Lozhkov, A., Wang, T., Karamcheti, S., Rush, A.M., Kiela, D., Cord, M., Sanh, V.: Obelics: An open web-scale filtered dataset of interleaved image-text documents (2023)

\bibitem{lester-etal-2021-power}
Lester, B., Al-Rfou, R., Constant, N.: The power of scale for parameter-efficient prompt tuning. In: Moens, M.F., Huang, X., Specia, L., Yih, S.W.t. (eds.) Proceedings of the 2021 Conference on Empirical Methods in Natural Language Processing. pp. 3045--3059. Association for Computational Linguistics, Online and Punta Cana, Dominican Republic (Nov 2021). \doi{10.18653/v1/2021.emnlp-main.243}, \url{https://aclanthology.org/2021.emnlp-main.243}

\bibitem{pmlr-v162-li22n}
Li, J., Li, D., Xiong, C., Hoi, S.: {BLIP}: Bootstrapping language-image pre-training for unified vision-language understanding and generation. In: Chaudhuri, K., Jegelka, S., Song, L., Szepesvari, C., Niu, G., Sabato, S. (eds.) Proceedings of the 39th International Conference on Machine Learning. Proceedings of Machine Learning Research, vol.~162, pp. 12888--12900. PMLR (17--23 Jul 2022), \url{https://proceedings.mlr.press/v162/li22n.html}

\bibitem{li2021align}
Li, J., Selvaraju, R., Gotmare, A., Joty, S., Xiong, C., Hoi, S.C.H.: Align before fuse: Vision and language representation learning with momentum distillation. Advances in neural information processing systems  \textbf{34},  9694--9705 (2021)

\bibitem{Li_2022_CVPR}
Li, L.H., Zhang, P., Zhang, H., Yang, J., Li, C., Zhong, Y., Wang, L., Yuan, L., Zhang, L., Hwang, J.N., Chang, K.W., Gao, J.: Grounded language-image pre-training. In: Proceedings of the IEEE/CVF Conference on Computer Vision and Pattern Recognition (CVPR). pp. 10965--10975 (June 2022)

\bibitem{li-liang-2021-prefix}
Li, X.L., Liang, P.: Prefix-tuning: Optimizing continuous prompts for generation. In: Zong, C., Xia, F., Li, W., Navigli, R. (eds.) Proceedings of the 59th Annual Meeting of the Association for Computational Linguistics and the 11th International Joint Conference on Natural Language Processing (Volume 1: Long Papers). pp. 4582--4597. Association for Computational Linguistics, Online (Aug 2021). \doi{10.18653/v1/2021.acl-long.353}, \url{https://aclanthology.org/2021.acl-long.353}

\bibitem{lin2023match}
Lin, W., Karlinsky, L., Shvetsova, N., Possegger, H., Kozinski, M., Panda, R., Feris, R., Kuehne, H., Bischof, H.: Match, expand and improve: Unsupervised finetuning for zero-shot action recognition with language knowledge. In: ICCV (2023)

\bibitem{lin2022learning}
Lin, X., Petroni, F., Bertasius, G., Rohrbach, M., Chang, S.F., Torresani, L.: Learning to recognize procedural activities with distant supervision. arXiv preprint arXiv:2201.10990  (2022)

\bibitem{liu2023mug}
Liu, et~al.: Mug-{STAN}: Adapting image-language pretrained models for general video. arXiv:2311.15075  (2023)

\bibitem{NEURIPS2022_00226294}
Liu, H., Li, C., Wu, Q., Lee, Y.J.: Visual instruction tuning. In: Advances in Neural Information Processing Systems. Curran Associates, Inc. (2023)

\bibitem{liu2023llava}
Liu, H., Li, C., Wu, Q., Lee, Y.J.: Visual instruction tuning. In: NeurIPS (2023)

\bibitem{liu2023revisiting}
Liu, R., Huang, J., Li, G., Feng, J., Wu, X., Li, T.H.: Revisiting temporal modeling for clip-based image-to-video knowledge transferring. In: Proceedings of the IEEE/CVF Conference on Computer Vision and Pattern Recognition. pp. 6555--6564 (2023)

\bibitem{LIU2023}
Liu, X., Zheng, Y., Du, Z., Ding, M., Qian, Y., Yang, Z., Tang, J.: Gpt understands, too. AI Open  (2023). \doi{https://doi.org/10.1016/j.aiopen.2023.08.012}, \url{https://www.sciencedirect.com/science/article/pii/S2666651023000141}

\bibitem{luo2022clip4clip}
Luo, H., Ji, L., Zhong, M., Chen, Y., Lei, W., Duan, N., Li, T.: Clip4clip: An empirical study of clip for end to end video clip retrieval and captioning. Neurocomputing  \textbf{508},  293--304 (2022)

\bibitem{menon2023visual}
Menon, S., Vondrick, C.: Visual classification via description from large language models. In: The Eleventh International Conference on Learning Representations (2023), \url{https://openreview.net/forum?id=jlAjNL8z5cs}

\bibitem{miech20rareact}
Miech, A., Alayrac, J.B., Laptev, I., Sivic, J., Zisserman, A.: Rareact: A video dataset of unusual interactions. arxiv:2008.01018  (2020)

\bibitem{owlvit}
Minderer, M., Gritsenko, A., Stone, A., Neumann, M., Weissenborn, D., Dosovitskiy, A., Mahendran, A., Arnab, A., Dehghani, M., Shen, Z., Wang, X., Zhai, X., Kipf, T., Houlsby, N.: Simple open-vocabulary object detection. In: Avidan, S., Brostow, G., Ciss{\'e}, M., Farinella, G.M., Hassner, T. (eds.) Computer Vision -- ECCV 2022. pp. 728--755. Springer Nature Switzerland, Cham (2022)

\bibitem{ni2022expanding}
Ni, B., Peng, H., Chen, M., Zhang, S., Meng, G., Fu, J., Xiang, S., Ling, H.: Expanding language-image pretrained models for general video recognition. In: European Conference on Computer Vision. pp. 1--18. Springer (2022)

\bibitem{pratt2023does}
Pratt, S., Covert, I., Liu, R., Farhadi, A.: What does a platypus look like? generating customized prompts for zero-shot image classification. In: Proceedings of the IEEE/CVF International Conference on Computer Vision. pp. 15691--15701 (2023)

\bibitem{radford2021learning}
Radford, A., Kim, J.W., Hallacy, C., Ramesh, A., Goh, G., Agarwal, S., Sastry, G., Askell, A., Mishkin, P., Clark, J., et~al.: Learning transferable visual models from natural language supervision. In: International conference on machine learning. pp. 8748--8763. PMLR (2021)

\bibitem{hanoonavificlip}
Rasheed, H., khattak, M.U., Maaz, M., Khan, S., Khan, F.S.: Finetuned clip models are efficient video learners. In: The IEEE/CVF Conference on Computer Vision and Pattern Recognition (2023)

\bibitem{vidla}
Rizve, et~al.: {VidLA}: Video-language alignment at scale. In: CVPR (2024)

\bibitem{Roth_2023_ICCV}
Roth, K., Kim, J.M., Koepke, A.S., Vinyals, O., Schmid, C., Akata, Z.: Waffling around for performance: Visual classification with random words and broad concepts. In: Proceedings of the IEEE/CVF International Conference on Computer Vision (ICCV). pp. 15746--15757 (October 2023)

\bibitem{schuhmann2022laion}
Schuhmann, C., Beaumont, R., Vencu, R., Gordon, C., Wightman, R., Cherti, M., Coombes, T., Katta, A., Mullis, C., Wortsman, M., et~al.: Laion-5b: An open large-scale dataset for training next generation image-text models. Advances in Neural Information Processing Systems  \textbf{35},  25278--25294 (2022)

\bibitem{schuhmann2021laion}
Schuhmann, C., Vencu, R., Beaumont, R., Kaczmarczyk, R., Mullis, C., Katta, A., Coombes, T., Jitsev, J., Komatsuzaki, A.: Laion-400m: Open dataset of clip-filtered 400 million image-text pairs. arXiv preprint arXiv:2111.02114  (2021)

\bibitem{sharma2018conceptual}
Sharma, P., Ding, N., Goodman, S., Soricut, R.: Conceptual captions: A cleaned, hypernymed, image alt-text dataset for automatic image captioning. In: Proceedings of the 56th Annual Meeting of the Association for Computational Linguistics (Volume 1: Long Papers). pp. 2556--2565 (2018)

\bibitem{shvetsova2023howtocaption}
Shvetsova, N., Kukleva, A., Hong, X., Rupprecht, C., Schiele, B., Kuehne, H.: Howtocaption: Prompting llms to transform video annotations at scale (2023)

\bibitem{singh2022flava}
Singh, A., Hu, R., Goswami, V., Couairon, G., Galuba, W., Rohrbach, M., Kiela, D.: Flava: A foundational language and vision alignment model. In: Proceedings of the IEEE/CVF Conference on Computer Vision and Pattern Recognition. pp. 15638--15650 (2022)

\bibitem{Swetha_xformer2024}
Sirnam, S., Yang, J., Neiman, T., Rizve, M.N., Tran, S., Yao, B., Chilimbi, T., Shah, M.: X-former: Unifying contrastive and reconstruction learning for mllms. In: Computer Vision -- ECCV 2024. Springer Nature Switzerland (2025)

\bibitem{sun2022dualcoop}
Sun, X., Hu, P., Saenko, K.: Dualcoop: Fast adaptation to multi-label recognition with limited annotations. Advances in Neural Information Processing Systems  \textbf{35},  30569--30582 (2022)

\bibitem{swetha2025implicitqa}
Swetha, S., Gupta, R., Kulkarni, P.P., Shatwell, D.G., Santiago, J.A.C., Siddiqui, N., Fioresi, J., Shah, M.: Implicitqa: Going beyond frames towards implicit video reasoning. arXiv preprint arXiv:2506.21742  (2025)

\bibitem{Swetha_icip}
Swetha, S., Kuehne, H., Rawat, Y.S., Shah, M.: Unsupervised discriminative embedding for sub-action learning in complex activities. In: 2021 IEEE International Conference on Image Processing (ICIP). pp. 2588--2592 (2021). \doi{10.1109/ICIP42928.2021.9506759}

\bibitem{swetha2025timelogic}
Swetha, S., Kuehne, H., Shah, M.: Timelogic: A temporal logic benchmark for video qa. arXiv preprint arXiv:2501.07214  (2025)

\bibitem{Swetha_2023_ICCV}
Swetha, S., Rizve, M.N., Shvetsova, N., Kuehne, H., Shah, M.: Preserving modality structure improves multi-modal learning. In: Proceedings of the IEEE/CVF International Conference on Computer Vision (ICCV). pp. 21993--22003 (October 2023)

\bibitem{10.1145/2812802}
Thomee, B., Shamma, D.A., Friedland, G., Elizalde, B., Ni, K., Poland, D., Borth, D., Li, L.J.: Yfcc100m: The new data in multimedia research. Commun. ACM  \textbf{59}(2),  64–73 (jan 2016). \doi{10.1145/2812802}, \url{https://doi.org/10.1145/2812802}

\bibitem{touvron2023llama}
Touvron, H., Martin, L., Stone, K., Albert, P., Almahairi, A., Babaei, Y., Bashlykov, N., Batra, S., Bhargava, P., Bhosale, S., et~al.: Llama 2: Open foundation and fine-tuned chat models. arXiv preprint arXiv:2307.09288  (2023)

\bibitem{wasim2023vita}
Wasim, S.T., Naseer, M., Khan, S., Khan, F.S., Shah, M.: Vita-clip: Video and text adaptive clip via multimodal prompting. In: Proceedings of the IEEE/CVF Conference on Computer Vision and Pattern Recognition. pp. 23034--23044 (2023)

\bibitem{weng2023open}
Weng, Z., Yang, X., Li, A., Wu, Z., Jiang, Y.G.: Open-{VCLIP}: Transforming {CLIP} to an open-vocabulary video model via interpolated weight optimization. In: Krause, A., Brunskill, E., Cho, K., Engelhardt, B., Sabato, S., Scarlett, J. (eds.) Proceedings of the 40th International Conference on Machine Learning. Proceedings of Machine Learning Research, vol.~202, pp. 36978--36989. PMLR (23--29 Jul 2023), \url{https://proceedings.mlr.press/v202/weng23b.html}

\bibitem{weng2023transforming}
Weng, Z., Yang, X., Li, A., Wu, Z., Jiang, Y.G.: Open-vclip: Transforming clip to an open-vocabulary video model via interpolated weight optimization. In: ICML (2023)

\bibitem{wortsman2022robust}
Wortsman, M., Ilharco, G., Kim, J.W., Li, M., Kornblith, S., Roelofs, R., Lopes, R.G., Hajishirzi, H., Farhadi, A., Namkoong, H., et~al.: Robust fine-tuning of zero-shot models. In: Proceedings of the IEEE/CVF Conference on Computer Vision and Pattern Recognition. pp. 7959--7971 (2022)

\bibitem{xu2023challenges}
Xu, Z., Zhu, Y., Deng, T., Mittal, A., Chen, Y., Wang, M., Favaro, P., Tighe, J., Modolo, D.: Challenges of zero-shot recognition with vision-language models: Granularity and correctness (2023)

\bibitem{xue2022clip}
Xue, H., Sun, Y., Liu, B., Fu, J., Song, R., Li, H., Luo, J.: Clip-vip: Adapting pre-trained image-text model to video-language representation alignment. arXiv preprint arXiv:2209.06430  (2022)

\bibitem{Yang_2023_CVPR}
Yang, Y., Panagopoulou, A., Zhou, S., Jin, D., Callison-Burch, C., Yatskar, M.: Language in a bottle: Language model guided concept bottlenecks for interpretable image classification. In: Proceedings of the IEEE/CVF Conference on Computer Vision and Pattern Recognition (CVPR). pp. 19187--19197 (June 2023)

\bibitem{yao2021filip}
Yao, L., Huang, R., Hou, L., Lu, G., Niu, M., Xu, H., Liang, X., Li, Z., Jiang, X., Xu, C.: Filip: Fine-grained interactive language-image pre-training. arXiv preprint arXiv:2111.07783  (2021)

\bibitem{yousaf2024videoprompter}
Yousaf, A., Naseer, M., Khan, S., Khan, F., Shah, M.: {VIDEOPROMPTER}: {AN} {ENSEMBLE} {OF} {FOUNDATIONAL} {MODELS} {FOR} {ZERO}-{SHOT} {VIDEO} {UNDERSTANDING} (2024), \url{https://openreview.net/forum?id=9F0xInGNBF}

\bibitem{yu2022coca}
Yu, J., Wang, Z., Vasudevan, V., Yeung, L., Seyedhosseini, M., Wu, Y.: Coca: Contrastive captioners are image-text foundation models. arXiv preprint arXiv:2205.01917  (2022)

\bibitem{llovi}
Zhang, C., Lu, T., Islam, M.M., Wang, Z., Yu, S., Bansal, M., Bertasius, G.: A simple llm framework for long-range video question-answering. CoRR  \textbf{abs/2312.17235} (2023), \url{https://doi.org/10.48550/arXiv.2312.17235}

\bibitem{zhao2023lavila}
Zhao, Y., Misra, I., Kr{\"a}henb{\"u}hl, P., Girdhar, R.: Learning video representations from large language models. In: CVPR (2023)

\bibitem{zhou2022learning}
Zhou, K., Yang, J., Loy, C.C., Liu, Z.: Learning to prompt for vision-language models. International Journal of Computer Vision  \textbf{130}(9),  2337--2348 (2022)

\bibitem{Zhu_2023_ICCV}
Zhu, X., Zhang, R., He, B., Guo, Z., Zeng, Z., Qin, Z., Zhang, S., Gao, P.: Pointclip v2: Prompting clip and gpt for powerful 3d open-world learning. In: Proceedings of the IEEE/CVF International Conference on Computer Vision (ICCV). pp. 2639--2650 (October 2023)

\end{thebibliography}

\end{document}